\documentclass{article}

\usepackage[preprint]{acl}
\usepackage{graphicx}%
\usepackage{multirow}%
\usepackage{amsmath,amssymb,amsfonts}%
\usepackage{amsthm}%
\usepackage[title]{appendix}%
\usepackage{xcolor}%
\usepackage{textcomp}%
\usepackage{booktabs}%

\usepackage{acronym}
\usepackage{multirow}
\usepackage{tabularx}
\usepackage{fancyvrb}
\usepackage{hyperref}
\usepackage{enumitem}
\usepackage{xcolor,colortbl}
\usepackage{siunitx}
\usepackage{orcidlink}

\setlist[enumerate]{noitemsep, nosep}
\setlist[itemize]{noitemsep, nosep}

\newtheoremstyle{colondef}
{3pt}   
{3pt}   
{\itshape}      
{}      
{\bfseries} 
{.}     
{ }     
{\thmname{#1}\thmnumber{ #2}\thmnote{: #3}}	

\theoremstyle{colondef}
\newtheorem{definition}{Definition}%
\newtheorem{finding}{Finding}
\newtheorem*{researchquestion}{RQ}

\newcommand{\librarylink}{\url{https://github.com/dimits-ts/apunim}}
\newcommand{\doclink}{\url{https://dimits-ts.github.io/apunim/}}
\newcommand{\explink}{\url{https://github.com/dimits-ts/aposteriori-unimodality}}

\AtBeginDocument{%
\mathchardef\Stdcomma=\mathcode`,
\mathcode`,="8000
}
\begingroup\lccode`~=`, \lowercase{\endgroup\def~}{\Stdcomma\,}

\newcommand\given[1][]{\:#1\vert\:}

\newcommand{\Dimorcid}{\orcidlink{0000-0002-5675-3939}}
\newcommand{\Johnorcid}{\orcidlink{0000-0001-9188-7425}}

\author{
	Dimitris Tsirmpas\Dimorcid, John Pavlopoulos\Johnorcid \\
	\small
	Athens University of Economics and Business, Greece (\texttt{\{dim.tsirmpas,annis\}@aueb.gr})\\
	\small
	Archimedes, Athena Research Center, Greece
}

\graphicspath{ {../graphs} {graphs}  }

\title{Are we chasing ghosts? Quantifying unattributable polarization, and attributing the rest to annotator groups}

\begin{document}

\maketitle

\begin{abstract}
	Standard agreement metrics often fail to capture systematic differences in opinion between minority and majority-group annotators, jeopardizing tasks such as hate speech and toxicity detection. Polarization has recently been proposed as a more robust way of distinguishing minor disagreements from systematic differences in opinion, but existing approaches do not provide practical tools for attributing it to specific annotator groups. We evaluate current methods and identify two major limitations in realistic settings: (1) the presence of ``inherent'' polarization that cannot be attributed to any known or latent groups, and (2) opposing polarization effects canceling each other out in aggregated annotations.
	To address these issues, we introduce a new metric that measures and tests the statistical significance of polarization attribution for annotator groups while avoiding these limitations, as well as an open-source Python library implementation, finding that no more than 20 annotators are needed per comment for reliable estimation. We apply our method to four subjective NLP datasets and find that gender and race consistently explain polarization patterns, while differences between annotator groups become stronger as the groups are further apart.
\end{abstract}

\section{Introduction}
\label{sec:intro}

\begin{figure}[ht]
	\centering
	\includegraphics[width=\columnwidth]{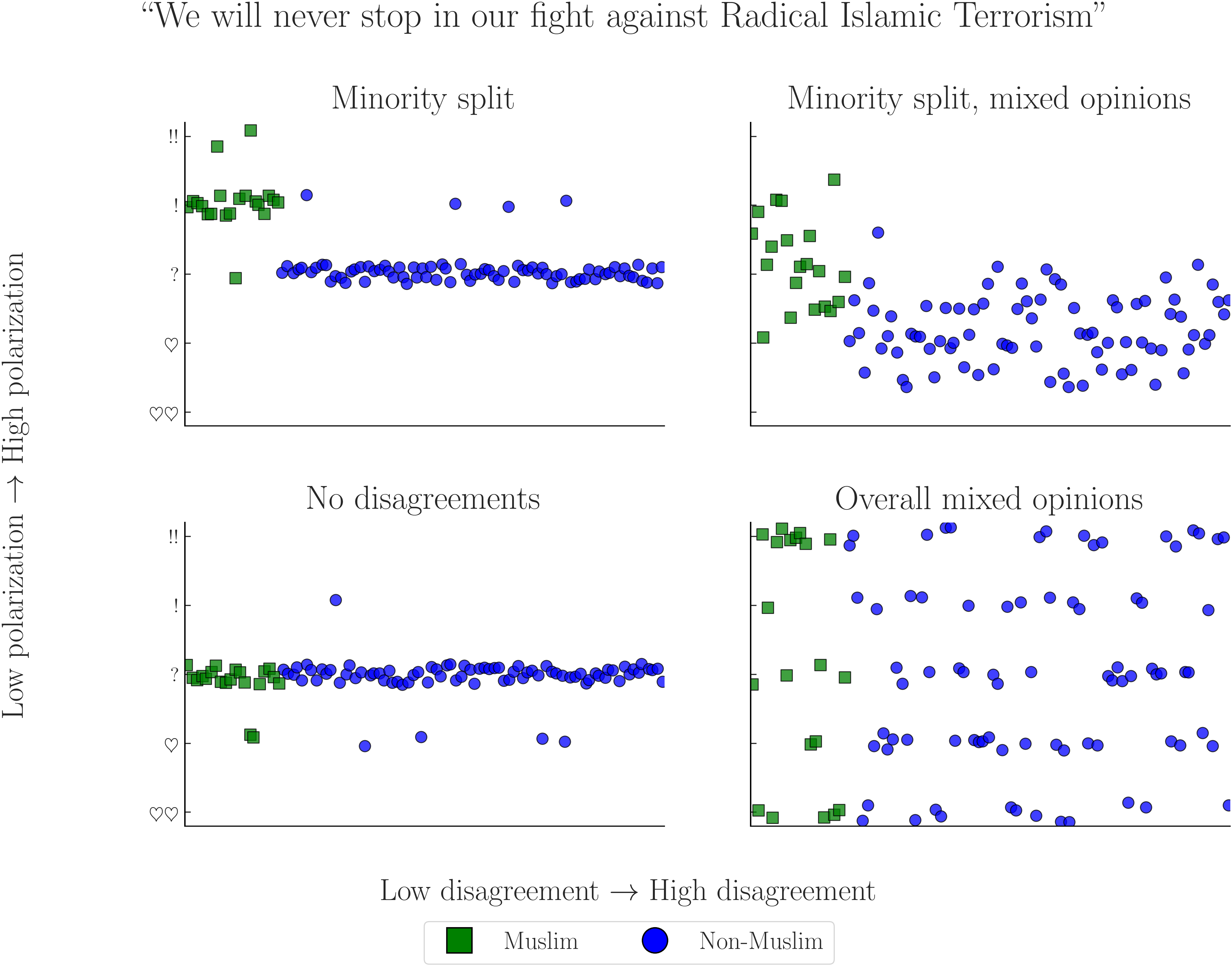}
	\caption{Simulated experiment exploring four scenarios of a hate speech detection task applied to a controversial statement by D. Trump (Jan. 2019). The variance in annotations across different annotators may cause traditional agreement metrics to ignore minority opinions, especially when opinions are mixed (top-right).}
	\label{fig:disagreement-vs-polarization}
\end{figure}

Annotations are essential for a wide range of tasks, especially in fields with subjective judgements, such as content moderation, toxicity, and hate speech detection. These tasks are especially important for training supervised learning models that remove or flag comments targeting vulnerable and minority groups in online spaces, where they are often vulnerable~\cite{un_hate_speech_targets, ucdavis_hate_speech}. However, in practice annotations are typically aggregated or filtered based on inter-annotator agreement or majority-vote~\cite{ivey-etal-2025-nutmeg, sokratis, petra_2022_handling, van_der_Velden}, which may at best limit the usefulness of the data, or at worst lead to biased and misconfigured systems~\cite{sap-etal-2019-risk, van_der_Velden}. Additionally, these practices may discard \emph{fundamental} disagreements between minority groups and the majority; for example, coded language may not be picked up by most annotators but only by ones belonging in the targeted minority~\cite{quaranto2022dog}, as shown in Fig.~\ref{fig:disagreement-vs-polarization}. Furthermore, majority-group annotators may themselves be biased against minority groups, e.g., toxicity models disproportionally marking \ac{aa} speech as ``toxic''~\cite{sap-etal-2019-risk}.

One alternative is using \emph{polarization}, which correlates better with human judgments and is much less sensitive to different group sizes \cite{Pavlopoulos2023, pavlopoulos-likas-2024}. While we can detect polarization, there is little work on how to detect whether it is \emph{actually caused by marginalized groups}. Our work thus tries to answer the following Research Question:

\begin{researchquestion}
	How much polarization can we practically attribute to annotator subgroups given existing datasets?
\end{researchquestion}

While prior work assumed that all polarization can be explained and attributed to some known or latent annotator characteristic, through the exhaustive creation of theoretical annotator groups our study discovers the existence of ``inherent polarization'', which cannot be attributed to \emph{any possible group} of annotators--known or unknown (Finding~\ref{finding:inherent}). This phenomenon could be attributed to the overall polarization of the data, or the limited number of annotations per item in many \ac{nlp} datasets. In any case, it establishes a hard limit on how much information we can gain from polarized items.

We also discover that polarization across comments has a \emph{direction}, preventing us from aggregating annotations on the dataset level to circumvent annotation sparsity (Finding~\ref{finding:conflicting}). Our analysis then pivots to estimating the number of annotators needed for robust polarization attribution, finding that no more than 20 annotators are typically needed, and that polarization reliability is asymptotically capped, even when both annotated items and annotations per item are abundant (with some caveats--see Finding~\ref{finding:ann-size}). These results suggest that current \ac{nlp} datasets may not have sufficient information for a thorough analysis of annotator behaviors, especially for highly subjective tasks.

In order to solve these issues we introduce a new metric--``\emph{apunim}''--that attributes polarization to specific annotator groups and develop a parametric test for statistical significance. Our metric is generalizable across any domain and modality (e.g., video/image classification) and, unlike previous approaches, enables \emph{quantitative} comparisons across datasets. By considering the entire set of comments in a dataset \emph{apunim} helps to identify genuine polarization and distinguish it from chance attribution, which is essential when analyzing large datasets without large annotator counts per item. Even in cases where unattributable polarization exists, \emph{apunim} is capable of detecting general patterns of polarization across multiple groups.

Having established our metric, we apply it to four \ac{nlp} datasets focused on Toxicity and Hate Speech detection (\S\ref{sec:results}), finding that annotator gender and race significantly impact annotation (Finding~\ref{finding:categorical}), and that fundamental disagreements grow more pronounced the further apart annotator groups are (e.g., irreligious vs. very religious annotators--Finding~\ref{finding:ordinal}). Finally, we release an open source Python (PyPi) library that implements our metric.\footnote{Library: \librarylink, Webpage: \doclink } The code for the experiments, graph creation, and analysis can be found on our project's repository.\footnote{Replication code: \explink}

\noindent\textbf{Warning: This publication includes examples of controversial and potentially harmful speech for the purposes of demonstration. These examples do not represent the views of the authors.}

\section{Background and related work}
\label{sec:related}

\subsection{Agreement}
\label{sec:related:agreement}

There are many popular metrics for measuring inter-annotator agreement such as Cohen's kappa, Krippendorff's alpha and Fleiss's kappa. All of these metrics attempt to distinguish genuine disagreement among annotators from noise (or ``random disagreement'') using statistical methods.

However, most cannot be generally used for comparing intergroup annotation agreement without extending them (e.g., computing pairwise Cohen's kappa for all groups), or placing constraints on when they can be used (e.g., no missing labels and same number of annotations per annotator for Fleiss's kappa). They have also been criticized for ignoring fundamental differences that can arise from different social, cultural and demographic backgrounds~\cite{Feinstein1990, Byrt1993, van_der_Velden, checco_2017}, for their inability to quantify and compare results across experiments~\cite{wong-etal-2021-cross}, for sensitivity towards skewed and imbalanced data~\citep{Feinstein1990}-- common for annotations of minority groups--and for fundamentally faulty assumptions around the concept of ``chance-correction''~\cite{checco_2017}.

Recent work has attempted to work around these limitations by using simpler statistical models~\cite{van_der_Velden}, adapting Cohen's kappa for subgroup agreement~\cite{wong-etal-2021-cross}, or creating a Bayesian model to explicitly differentiate between subgroup disagreement and noise~\cite{ivey-etal-2025-nutmeg}. While these extensions do improve the performance of these metrics, they are typically less explainable and intuitive.

\subsection{Polarization}
\label{sec:related:polarization}

There are multiple competing formulations for polarization. One of these approaches~\cite{akhtar_2019} utilizes $\chi^2$ to estimate in-group vs out-group variance, but still necessitates traditional agreement metrics for analysis.
Others attempt to solve this problem by framing disagreement as a cluster identification problem, applying either parametric~\cite{checco_2017}, or non-parametric~\cite{pavlopoulos-likas-2024, Mignemi2024} approaches to the annotation histograms. 

For our work, we use a non-parametric polarization metric, \ac{ndfu}, which has been shown to correlate better with human judgment of disagreement than established metrics \citep{pavlopoulos-likas-2024}. \ac{ndfu} takes values in the range $[0,1]$, where values close to $0$ indicate little or no polarization (i.e., opinions are concentrated around a single mode), while values close to $1$ indicate strong polarization with multiple distinct peaks. The metric is defined as follows:

\begin{equation}
	\label{eq:ndfu}
	\begin{aligned}
		\mathrm{nDFU}(X)
		&=
		\frac{
			\displaystyle
			\max_{i \neq m}
			\left(
			f_i -
			\begin{cases}
				f_{i-1}, & \text{if } i > m \\
				f_{i+1}, & \text{if } i < m
			\end{cases}
			\right)
		}{
			f_m
		} \\
		& m = \arg\max_i f_i
	\end{aligned}
\end{equation}

\noindent where, $X$ denotes the set of annotations, $f_i$ the relative frequency of the $i$-th annotation (e.g., the frequency of \texttt{2} in a scale of 1--5), and $m$ the histogram peak. Intuitively, \ac{ndfu} measures how prominent secondary peaks are compared to the main peak (see Fig.~\ref{fig:ndfu_combined} for an example).

\subsection{Attributing polarization to groups}
\label{sec:related:attribution}
While many studies study the effect of subgroups on annotation agreement~\cite{goyal2022your, binns2017like, giorgi2025human, al2020identifying, sachdeva2022assessing}, attributing polarization to subgroups is a relatively new idea. Under the name \emph{aposteriori unimodality}~\cite{pavlopoulos-likas-2024}, the original idea is that if polarization in a single item is positive, but zero for all annotator subgroups, then those subgroups are ``causing'' the exhibited polarization. Though promising, this mechanism operates on an edge-case scenario where \emph{all} polarization is explained by a single split. Additionally, getting polarization estimates from few annotations per item is extremely noisy, which is especially problematic when the output is binary (i.e., an item either is or is not polarizing) instead of a quantifiable value.\footnote{For example, when presented with noisy estimates of a metric in the  $[0,1]$ range, we would ideally want to distinguish between cases where attribution is close to $0.001$, and cases where attribution is $0.49$.} 

Another approach~\citep{sokratis} attempts to first filter out non-substantial disagreements using polarization, and then analyzes differences between LLMs and annotator subgroups through traditional agreement metrics. This analysis however is limited by the inherent issues of using traditional agreement metrics for minority groups (\S\ref{sec:related:agreement}), and the authors prioritize comparisons between human and LLM annotators instead of a deeper analysis of minority groups.

\section{Is it possible to attribute all polarization to annotator groups?}
\label{sec:inherent}

\subsection{Problem Formulation}
\label{sec:inherent:formulation}

Let $d = \{c_1, c_2, \ldots\}$ be a dataset composed of multiple annotated comments.\footnote{These items could be of any modality, such as comments in a discussion, images, videos or survey questions, even though we consider comments in this study.} Each comment $c$ is assigned multiple annotators,  which creates the annotation multi-set $A(c)=  \{(a_1, g_1), (a_2, g_2), \ldots \}$, where $a_i$ is a single annotation (e.g., \texttt{2} in a LIKERT scale of 1--5). As an example, the annotations for a comment in a sentiment analysis task where we group the annotators by gender, would be formulated as:
\begin{multline*}
	A(\text{``Could be better, could be worse''})\\
	= \left\{\begin{array}{l}
		(\text{+}, F), (\text{--}, M), (\text{+}, F), \ldots
	\end{array}\right\}
\end{multline*}

\begin{figure}[t]
	\centering
	\includegraphics[width=\columnwidth]{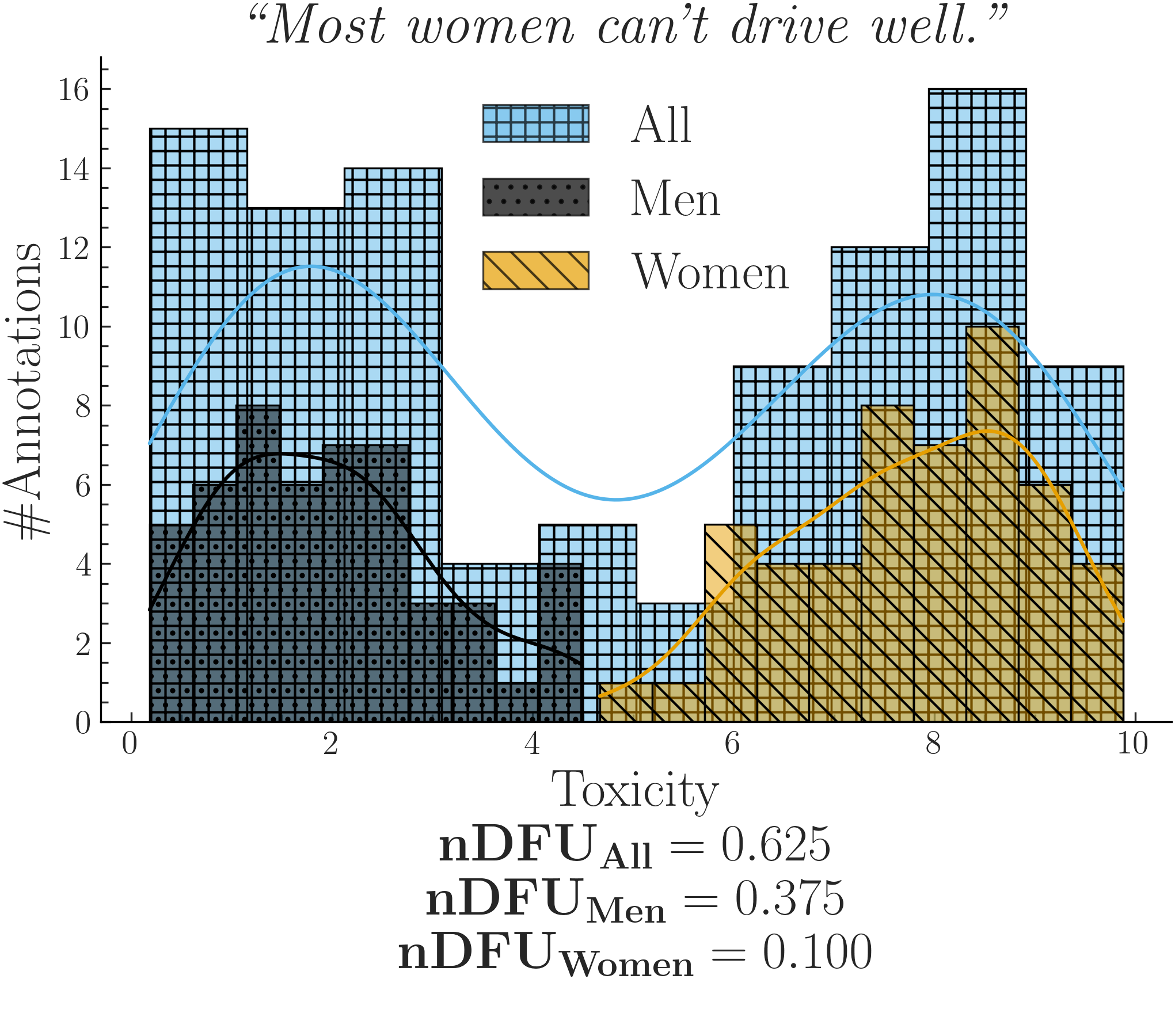}
	\caption{Simulated experiment describing a polarizing comment where male and female annotators agree between themselves, but disagree with the opposite gender.}
	\label{fig:ndfu_combined}
\end{figure}

Intuitively, a group partially explains the polarization of a comment $c$ when the annotations of that group $A(c \given g) = \{a(c, g') \in A(c) \given g'=g\}$ exhibit higher polarization compared to the full set of annotations $A(c)$. Consider the case where a misogynistic comment is annotated for toxicity by male and female annotators shown in Fig.~\ref{fig:ndfu_combined}.\footnote{All simulated experiments in this study were implemented by generating random annotations following the normal distribution with differing means and std dev--see \explink.} The annotations are generally polarized ($nDFU_{all}=0.62$); but both female ($nDFU_{women}=0.10$) and male ($nDFU_{men}=0.37$) annotators exhibit low polarization between their respective groups. This pattern suggests that the overall polarization is partially caused by substantive disagreements between male and female annotators.\footnote{\label{foot:iterative}The continued polarization observed within the male annotators may suggest that a specific subgroup of men could also contribute to this polarization.}

\subsection{Datasets}
\label{sec:inherent:datasets}

\begin{table}[t]
	\centering
	\small
	\begin{tabular}{lrrrr}
		\toprule
		\textbf{Dataset} & \textbf{\#Com} & \textbf{\#Ann} & \textbf{\#Dims} & \textbf{Task} \\
		\midrule
		Kumar
		& 106{,}035 & 5   & 10 & Tox \\
		
		Sap 
		& 626       & 5--6   & 3  & HS \\
		
		DICES-350 
		& 72{,}103  & 70--76 & 4  & HS \\
		
		DICES-990
		& 43{,}050  & 123    & 4  & HS \\
		\bottomrule
	\end{tabular}
	\caption{Datasets used in this study. \textit{\#Ann} refers to the number of annotations per comment (see App.~\ref{sec:app:ann_details}), and \textit{\#Dims} to sociodemographic dimensions (e.g., age, gender). Tasks are either toxicity (Tox) or hate speech detection (HS).}
	\label{tab:datasets}
\end{table}

We use four datasets from current \ac{nlp} research that include sociodemographic information at the level of individual annotators; specifically, the ``Kumar'' \cite{kumar-et-al-2021} and ``Sap'' \cite{sap-etal-2022-annotators} datasets, which are concerned with Toxicity and hate speech detection respectively, and the DICES-350 and DICES-990 datasets~\cite{dices}, which focus on LLM response safety.\footnote{For the latter, we consider only labels relating to racism (\texttt{Q3\_bias\_overall}), thereby transforming the task to hate speech detection.} While the toxicity and hate speech detection tasks are not equivalent, they share substantial overlap, as both require annotators to assess the presence and severity of harmful or abusive language. Thus, we expect the sources of annotator disagreement and bias that arise in these tasks will be broadly similar.

A brief overview of dataset characteristics is provided in Table~\ref{tab:datasets}. The Kumar dataset stands out for its large size, featuring more than 100{,}000 annotated comments, as well as 10 distinct sociodemographic dimensions. In contrast, the DICES datasets contain fewer annotated comments but involve more than an order of magnitude more annotators per comment than the other datasets. The Kumar and Sap datasets contain a large number of polarizing items, unlike the DICES datasets, where most items are unpolarized (App.~\ref{sec:app:ann_details}; Fig.~\ref{fig:dataset-polarization}).\footnote{\label{foot:dices-pol}This systematic discrepancy could be attributed to the DICES datasets being composed of LLM-generated dialogues with already-aligned models tuned to avoid undafe speech. Even if such speech exists, it is likely that it is much easier to distinguish than similar comments in actual human discussions.}

\subsection{Inherent polarization}
\label{sec:inherent:inherent}

Inherent polarization refers to disagreement that cannot be attributed to \emph{any specific} grouping of annotators; that is, polarization that exists no matter how we split the annotators in any arbitrary group. Prior work~\cite{pavlopoulos-likas-2024} attributed polarization to annotator groups only when the observed polarization for each individual group became exactly zero (e.g. $nDFU(all) > nDFU(men)=nDFU(women) = 0$). The presence of inherent polarization would invalidate this theory, since there would be no possible group or combination of groups $g \textit{ s.t. } nDFU(g) = 0$.

\begin{definition}[Inherent polarization]
	\label{def:inherent}
	Polarization that cannot be explained by any known or latent annotator grouping.
\end{definition}

Formally, the inherent polarization for a single comment $c$ is given by:
\begin{equation}
	\label{eq:inherent}
	Pol_{inh}(c) = \min_{i} \{\textit{nDFU}(\tilde{r}_i (A(c)))\}	
\end{equation}
\noindent where $\tilde{r}$ is the random partition operator.

\subsection{Experimental setup}
\label{sec:inherent:experimental}

We systematically explore the space of annotator groups by generating random partitions for each comment. When considered exhaustively, these theoretical, random groups include both observed attributes (e.g., men and women) and latent or unobserved factors (e.g., annotators may be less strict if they had lunch prior to annotation). Since \ac{ndfu} operates only when a group has at least three annotators, this yields $\sum_{k=3}^{n} \binom{n}{k}$ possible groups, over which we test whether \emph{any} exhibit intra-group polarization:

\begin{equation}
	\label{eq:exhaustive}
	Pol_{inh}(c) =
	\min_{S \subseteq A(c),\, |S| \ge 3}
	\left\{
	\textit{nDFU}(S)
	\right\}
\end{equation}

Since enumerating all subsets is computationally intractable for large annotator counts, we approximate this quantity via Monte Carlo sampling when $\lvert A(c) \rvert \ge 10$. Specifically, we draw $1000$ random partitions of the annotators into groups of random sizes (each group containing at least three annotators), and compute:

\begin{equation}
	\label{eq:random}
	Pol_{inh}^{MC}(c) =
	\min_{j=1}^{1000}
	\left\{
	\textit{nDFU}\!\left(\tilde{r}_j(A(c))\right)
	\right\}
\end{equation}

Because this Monte Carlo approximation considers only a strict subset of the total possible configurations, the approximation provides a lower bound:  
\begin{equation}
	\label{eq:lower-bound}
	Pol_{inh}^{MC}(c) \ge Pol_{inh}(c) \forall c
\end{equation}

\subsection{Does it exist?}
\label{sec:inherent:results}

\begin{figure}[ht]
	\includegraphics[width=\columnwidth]{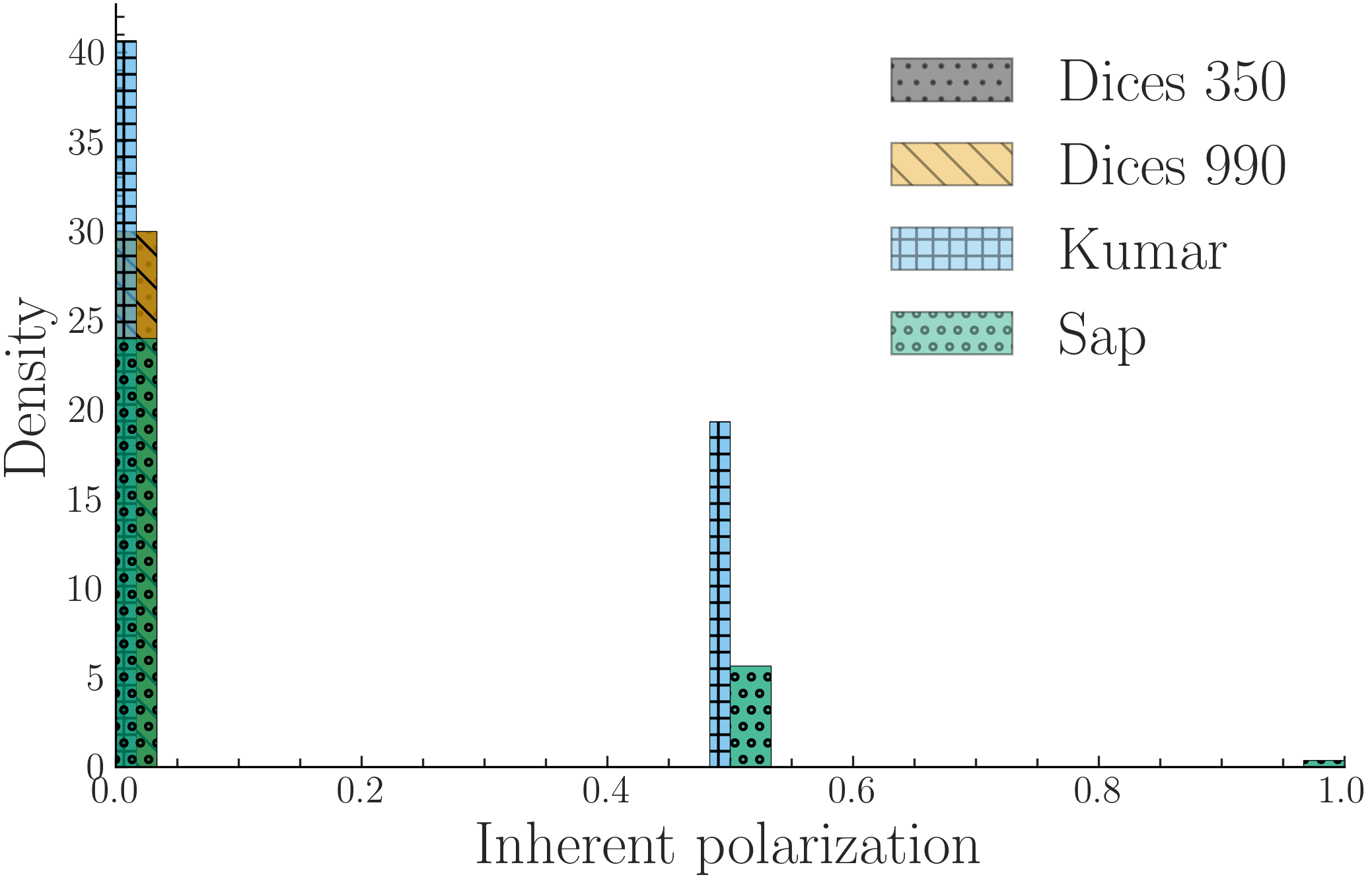}
	\caption{Inherent polarization across our datasets (\S\ref{sec:inherent:datasets}). Contrary to assumptions made in prior work, inherent polarization is commonly non-zero and can even reach $nDFU = 1$ in datasets with low annotator counts.}
	\label{fig:apriori}
\end{figure}

We compute the inherent polarization for each comment across the four datasets. For Sap and Kumar, we use the exhaustive formulation (Eq.~\ref{eq:exhaustive}), whereas for the two DICES datasets we rely on the Monte Carlo approximation (Eq.~\ref{eq:random}) due to the annotator counts making the exhaustive formulation intractable. Fig.~\ref{fig:apriori} shows that in all DICES comments there exists at least one group that fully accounts for the observed polarization.\footnote{$Pol_{inh}^{MC}(c)=0 \Rightarrow Pol_{inh}(c)=0$ due to Eq.~\ref{eq:lower-bound}. The same logic applies if we tried to increase the number of random partitions in Eq.~\ref{eq:random}. } Interestingly, this observation does not hold for the Kumar and Sap datasets, despite evaluating all possible annotator groups exhaustively. Further experiments show that repeating the experiment for subsets of (5--50) annotators sampled ten times with replacement also lead to zero inherent polarization in the DICES datasets.

We hypothesize that this discrepancy is caused by the existence of a latent group which is not represented in the annotator population i.e., although a latent group could in principle explain the observed polarization, such a group may be \emph{not represented when only a few annotators are available}. This hypothesis explains why the Kumar and Sap datasets, which feature low annotator counts, feature comments exhibiting non-zero inherent polarization. In contrast, the DICES datasets are much less polarized (see App.~\ref{sec:app:ann_details}; Fig.~\ref{fig:dataset-polarization} and Footnote~\ref{foot:dices-pol}); hence, all polarization can be explained even when a low annotator count is sampled for each comment.

\begin{finding}
	\label{finding:inherent}
	Comments may exhibit a degree of inherent polarization that cannot be attributed to any known or latent annotator grouping given the annotators present in the dataset.
\end{finding}

This finding suggests that current \ac{nlp} datasets may be \emph{lacking fundamental information} for analyzing annotator behavior, especially in very subjective and sensitive tasks. Even though the Kumar dataset employs an impressive $17{,}280$ annotators in total, the lack of annotations per comment results in a large section of these comments featuring unattributable polarization.

\section{How do we attribute the rest?}
\label{sec:apunim}

\subsection{From inherent to apriori polarization}
\label{sec:apunim:group-vs-global}

In \S\ref{sec:inherent:formulation}, we explained how our methodology needs to compare the polarization of $A(c)$ and $A(c \given g)$. Direct comparisons are not advisable, since we would be comparing statistics between a set and its subset--in this case, standard practice dictates comparing the subset with its complement; i.e., the in-group vs. the out-group instead of the in-group vs. all the annotations. However, this also would not work; if we apply the subset-vs-complement method in the example of Fig.~\ref{fig:ndfu_combined}, we would end up subtracting the \acp{ndfu} of men and women, which are similar as they are both unimodal distributions, falsely concluding that there is no polarization present.

We instead compare the observed polarization of $A(c)$ with its inherent polarization, following the methodology introduced in \S\ref{sec:inherent:inherent} with two key distinctions. (1) Instead of using randomly sized splits, we create random stratified splits.\footnote{I.e., if we have $100$ annotations, $80$ of which are made by male annotators and $20$ by female annotators, we create random partitions of sizes $80$ and $20$.} (2) Instead of finding the minimum possible polarization exhibited by any known or latent group, we calculate the \emph{expected} (average). Therefore, we do not test whether the group is polarized against the whole; nor whether the group explains \emph{some polarization} (as would be the case if we used Eq.~\ref{eq:inherent}), but rather whether the group explains polarization \emph{more than other groups}, which is a much stricter test.\footnote{Indeed, by definition, all groups exhibit equal or larger polarization than the inherent pol. value.} 

\begin{definition}[Apriori polarization]
	\label{def:apriori}
	The expected polarization of a comment's annotations.
\end{definition}

Specifically, we define apriori polarization for a comment $c$ as:
\begin{equation}
	\label{eq:apriori}
	Pol_{apr}(c) = \frac{1}{t} \sum_{i=1}^t  \textit{nDFU}(\tilde{r}_i (A(c)))
\end{equation}
\noindent where $\tilde{r}$ is the random partition operator (similar to Eq.~\ref{eq:inherent}, but with matching group sizes).

\subsection{Polarization is directional}
\label{sec:apunim:conflicting}

\begin{figure*}[ht]
	\centering
	\includegraphics[width=0.32\linewidth]{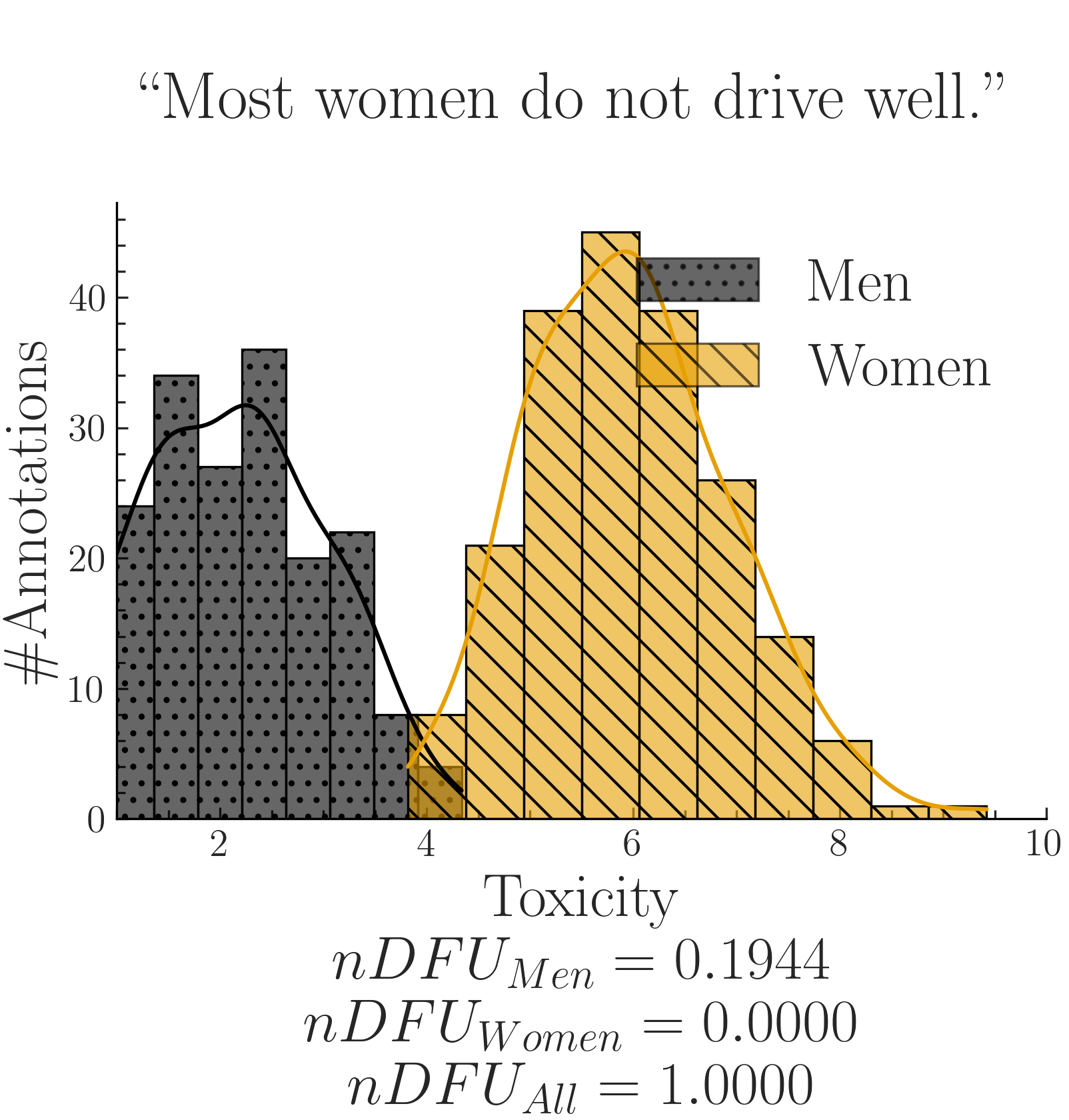}
	\includegraphics[width=0.32\linewidth]{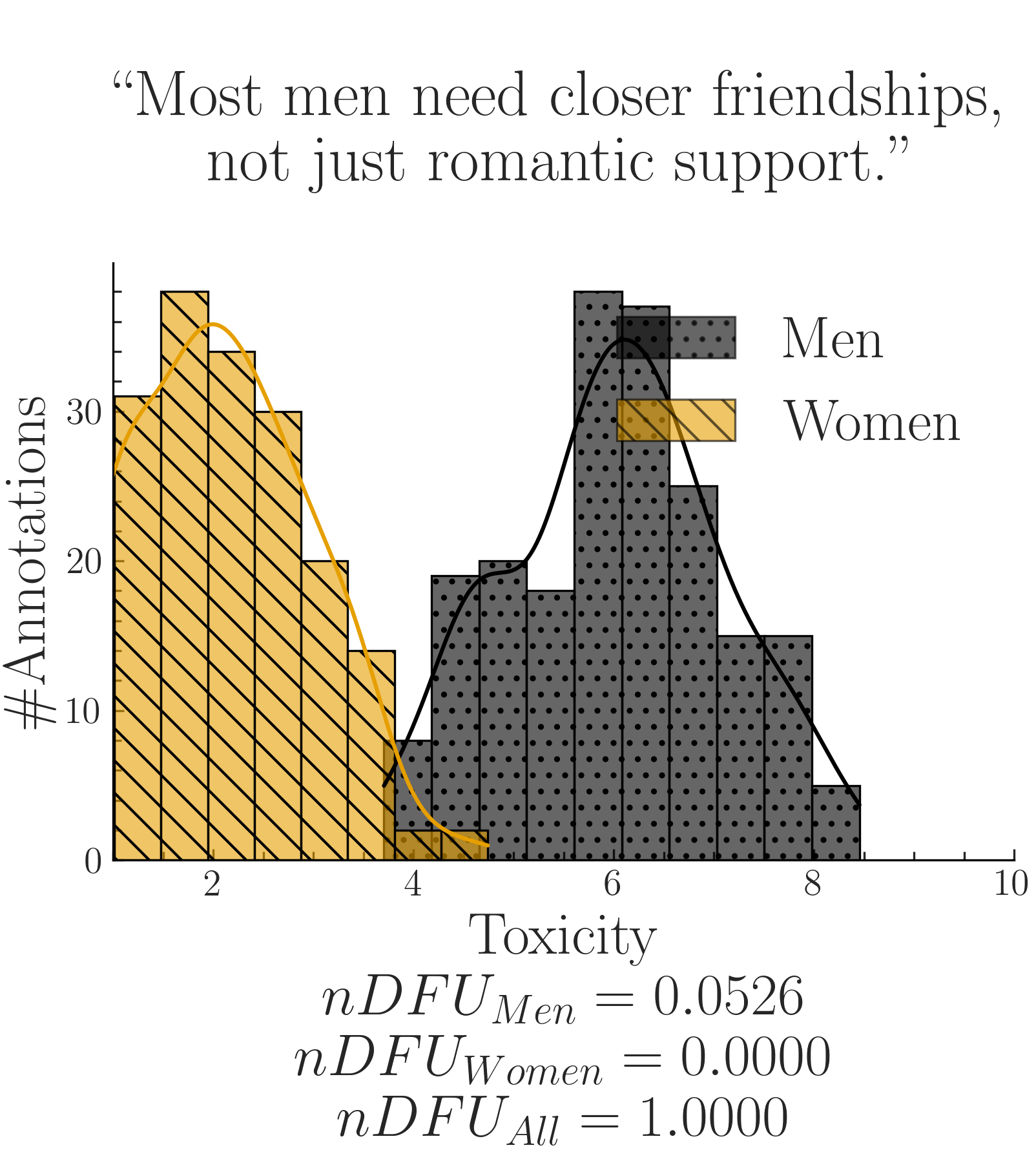}
	\includegraphics[width=0.32\linewidth]{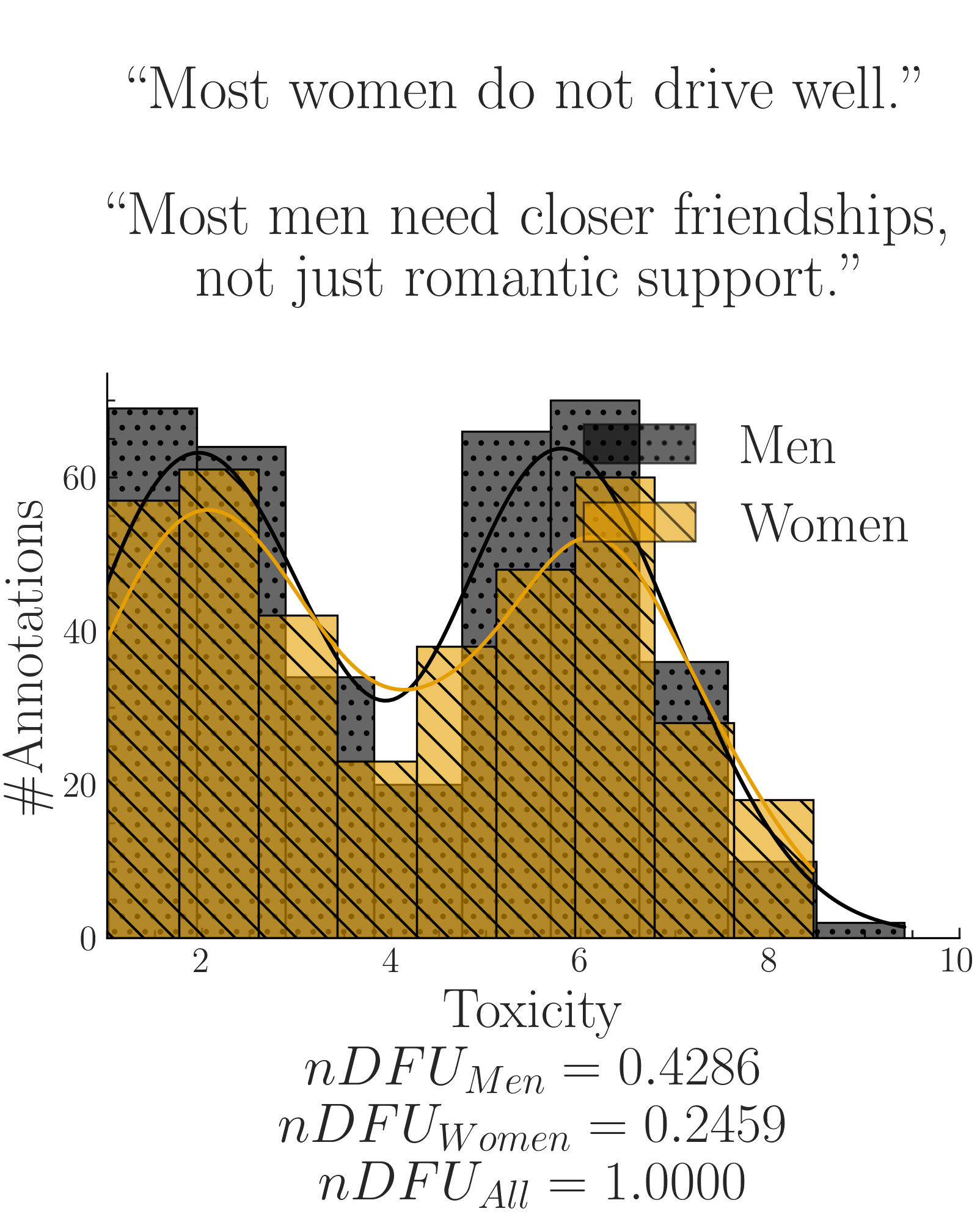}
	\caption{A synthetic experiment simulating a polarizing discussion with two comments where annotators disagree about which one is toxic. When we aggregate the comments, we unintentionally change the comparison: instead of comparing two unimodal distributions (each comment viewed separately) with two bimodal distributions (their aggregated annotations), we end up comparing two bimodal distributions, obscuring the source of the polarization (polarization explained by gender, but not shown).}
	\label{fig:ndfu_discussion}
\end{figure*}

Previous work~\cite{pavlopoulos-likas-2024} operated on the comment level. In practice however, this approach is inherently noisy due to a lack of information (i.e., limited annotations) relating to each individual comment. Many datasets in \ac{nlp} are annotated by only 3--5 annotators per comment, which is especially problematic when we want to analyze patterns in annotator subgroups; in a best-case scenario of 5 annotators split evenly between two groups, we would get a 3-2 split. This issue is especially prevalent for polarization, since \emph{\ac{ndfu} is undefined for groups of two} (and arguably marginally useful for groups of three), since it requires at least two histogram peaks to function (see Eq.~\ref{eq:ndfu}).

Thus, we consider whether simply aggregating the annotations on the dataset level would solve issues in data sparsity. Fig.~\ref{fig:ndfu_discussion} shows an experiment where we combine annotations across two comments. We observe that, should the two comments be polarized for the opposite reason, the opposing polarization effects might balance each other out, leading to a false negative. Therefore, we should never aggregate annotations across comments.\footnote{A similar assumption was followed in early parametric approaches, which assumed different parameters for each comment~\cite{checco_2017}.}

\begin{finding}
	\label{finding:conflicting}
	Polarization has a direction, which is unique for each annotated item, and may compromise cross-comment annotation aggregation.
\end{finding}

\subsection{Expanding to the whole dataset}
\label{sec:apunim:dataset}

We can estimate how much polarization is generally present in the dataset by simply averaging the \textit{apriori} polarization of its comments. Intuitively, a high apriori polarization score means that much of the polarization in the dataset is likely not driven by any single group.  
\begin{equation}
	\label{eq:apriori-dataset}
	Pol_{apr}^{d} = \frac{1}{\lvert d \rvert} \sum_{c \in d} Pol_{apr}(c)
\end{equation}

Which we will compare with the average observed polarization of all comments:
\begin{equation}
	\label{eq:observed}
	Pol_{obs}^{d}(g) = \frac{1}{\lvert d \rvert} \sum_{c \in d}  \textit{nDFU}(A(c \given g))
\end{equation}

We can now compare the polarization that is attributed to a single annotator group (Equation~\ref{eq:observed}) to a prior one (Equation~\ref{eq:apriori-dataset}). We define our normalized metric, \emph{apunim}, as:

\begin{equation}
	\label{eq:apunim}
	apunim(d, g) = \frac{Pol_{obs}^{d}(g) - Pol_{apr}^{d}}{1 - Pol_{apr}^{d}}
\end{equation}

\begin{definition}[Apunim]
	\label{def:apunim}
	The degree to which an annotator group contributes to overall polarization in a set of comments.
\end{definition}

Since $nDFU \rightarrow [0,1] \Rightarrow apunim \rightarrow [-1,1]$. The interpretation of this metric boils down to three scenarios:
\begin{itemize}
	\item$apunim(\theta) \approx 0$: The examined group does not meaningfully disagree with the other annotators.
	
	\item $apunim(\theta) > 0$: The examined group has fundamental disagreements with the other annotators. 
	
	\item $apunim(\theta) < 0$: This occurs when intragroup polarization is higher than apriori polarization; indicating that annotators of that group are polarized between themselves.
\end{itemize}

Because \textit{apunim} is fundamentally designed to quantify polarization, we exclude comments that exhibit no measurable polarization ($nDFU(c)=0$). Furthermore, to enhance computational efficiency, we also exclude comments annotated exclusively by a single group (e.g., comments annotated only by male or only by female annotators when considering gender).

\subsection{Statistical significance}
\label{sec:apunim:statistical}

Since annotation histograms may be noisy, we additionally assess whether the observed apunim value could arise by chance. To assess statistical significance, we construct a null distribution of apunim values from the randomized partitions and compare the observed apunim against this distribution using a one-sample Student-\(t\) test. Thus, the null hypothesis states that the observed polarization is indistinguishable from polarization induced by random annotator assignments:
\[
H_0:\ 
\bar{\textit{apunim}}_{\mathrm{rand}}(d,g)
=
\textit{apunim}(d,g)
\]
\[
H_a:\ 
\bar{\textit{apunim}}_{\mathrm{rand}}(d,g)
\neq
\textit{apunim}(d,g)
\]
\noindent where 
\(\bar{\textit{apunim}}_{\mathrm{rand}}(d,g)\) denotes the average apunim values obtained from $t$ randomized annotator partitions. For a detailed explanation of the p-value computation and the assumptions underlying the test, refer to App.~\ref{app:pvalue}.

\subsection{How many annotators do we need?}
\label{sec:apunim:ann-size}

\begin{figure}[t]
	\includegraphics[width=\columnwidth]{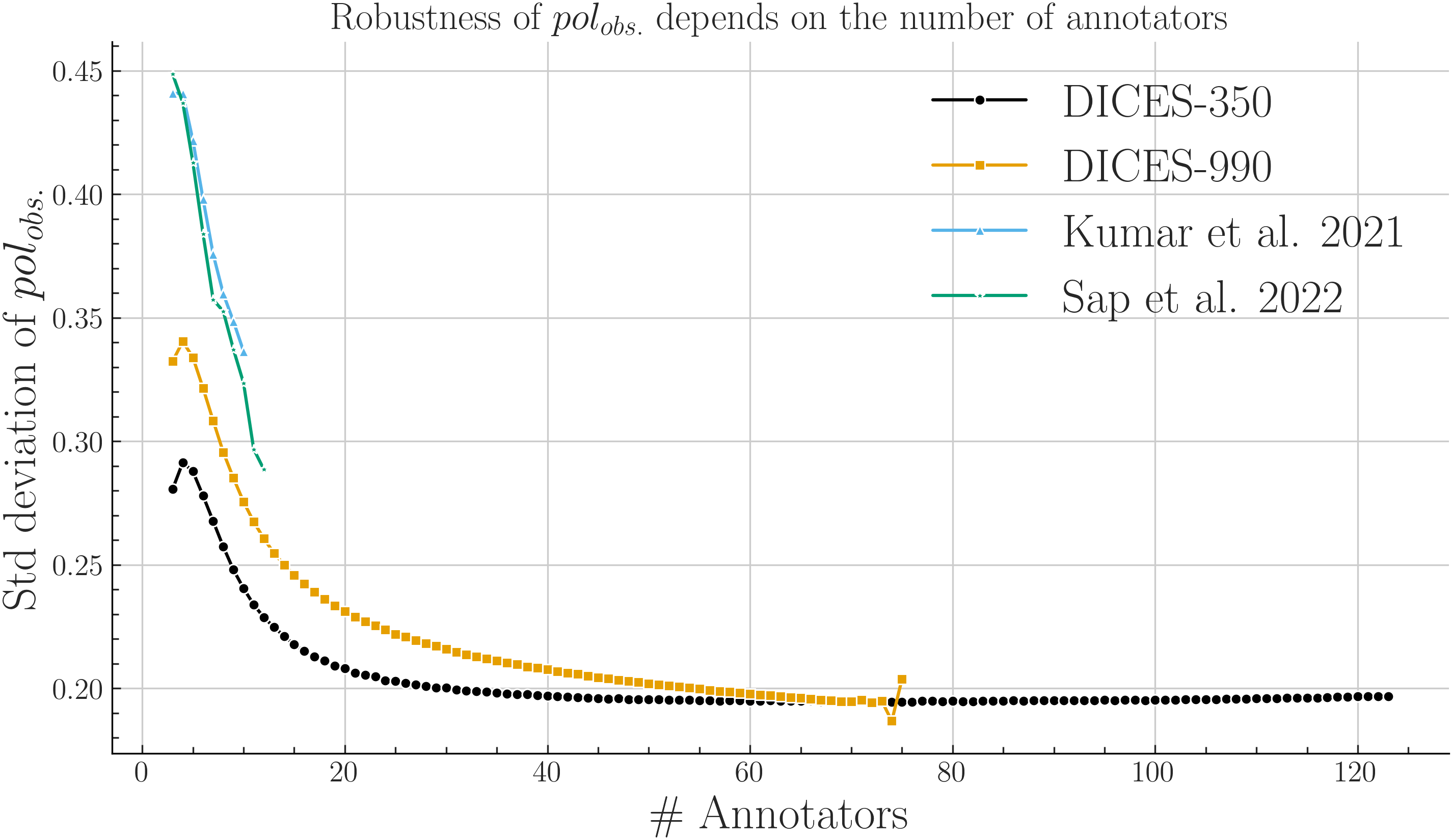}
	\caption{Standard deviation of 30 randomly sampled (with replacement) $pol_{obs}$ values (Eq.~\ref{eq:observed}). The x-axis shows the sample size (number of annotators), starting from 3 and increasing one at a time, up to the largest annotator count for which a sufficient number of items exists (App.~\ref{sec:app:ann_details}).}
	\label{fig:std_error}
\end{figure}

Fig.~\ref{fig:std_error} shows the impact of annotator count on the reliability of polarization estimation. Each dataset exhibits a unique standard deviation, even when the annotator count is fixed--which aligns broadly with the level of inherent polarization within that specific dataset (App.~\ref{sec:app:ann_details}). We observe a clear, non-linear, inverse relationship between the number of annotators and the standard error across all datasets, which can be attributed to better histogram estimation employed by \ac{ndfu} (\S\ref{sec:apunim:group-vs-global}) when supplied with greater point density for each group, although the rate of improvement becomes increasingly marginal after 20 annotators. Interestingly, some variability in polarization persists even with an extraordinary amount of annotators, which we hypothesize to be caused by inherent noise in the dataset.\footnote{To our knowledge, no datasets other than DICES contain a substantial number of annotators per item. Therefore, this finding should be interpreted as the strongest evidence currently available for this question, although we cannot assess whether it generalizes to other datasets.}

\begin{finding}
	\label{finding:ann-size}
	To achieve a reasonably reliable estimation of polarization attribution, a maximum of 20 annotators per item is sufficient. However, there is a definite ceiling to the reliability of polarization, irrespective of annotator count.
\end{finding}

This finding reinforces the fact that most \ac{nlp} datasets are lacking fine-grained information, as discussed in \S\ref{sec:inherent:results}. However, the indication that only $20$ annotators are needed for robust estimations of polarization is encouraging; it is much easier to scale up to this number of annotators compared to creating specializing datasets such as DICES which feature up to 100 annotators. Indeed, our formulation enables us to attribute polarization even in cases with very low annotator counts, as shown in the next section.

\section{Attributing real-life polarization to human groups}
\label{sec:results}

\subsection{Experimental Setup}
\label{sec:results:experimental}

Assuming a confidence level of $\alpha=0.05$ we test whether any of the provided sociodemographic dimensions can partially explain polarization in the toxicity / hate speech annotations, as shown in Table~\ref{tab:datasets}.\footnote{We sample 60{,}000 comments from the Kumar dataset due to memory constraints.} Note that some groups are not included because there were no polarized items annotated by more than one annotator of these groups. See App.~\ref{app:replication} for details on other parameters and execution times, and App.~\ref{app:full} for the full results.

\subsection{Which factors contribute to polarization?}
\label{ssec:results:factors}

\begin{table}[t]
	\small
	\centering
	
	\begin{tabularx}{\columnwidth}{X r r}
		\toprule
		\textbf{Group} & \textbf{apunim} & \textbf{support} \\
		\midrule
		
		\multicolumn{3}{c}{\textbf{DICES-350}} \\
		\midrule
		
		\rowcolor{gray!15}
		Race=AfricanAmerican & -0.043 & 9077 \\
		Race=Asian & 0.066 & 8138 \\
		
		\midrule
		\multicolumn{3}{c}{\textbf{DICES-990}} \\
		\midrule
		
		\rowcolor{gray!15}
		Age=GenX+ & -0.046 & 14384 \\
		Age=GenZ & 0.063 & 13741 \\
		
		\rowcolor{gray!15}
		Edu=High school and below & -0.075 & 7419 \\
		
		Gender=Man & 0.028 & 32494 \\
		
		\rowcolor{gray!15}
		Gender=Woman & -0.025 & 33471 \\
		
		Race=AfricanAmerican & 0.091 & 5810 \\
		
		\rowcolor{gray!15}
		Race=Latino & -0.133 & 6610 \\
		Race=Other & 0.075 & 24321 \\
		
		\rowcolor{gray!15}
		Race=White & -0.120 & 11392 \\
		
		\midrule
		\multicolumn{3}{c}{\textbf{Sap}} \\
		\midrule
		
		Gender=Man & 0.091 & 1439 \\
		
		\rowcolor{gray!15}
		Gender=Woman & -0.244 & 877 \\
		
		\midrule
		\multicolumn{3}{c}{\textbf{Kumar}} \\
		\midrule
		Race=AfricanAmerican & $0.2072$ & 356 \\
		\rowcolor{gray!15} ToxTarget=False & $-0.0017$ & 2199 \\
		ToxTarget=True & $0.0329$ & 1176 \\
		\rowcolor{gray!15} Parent=No & $-0.0591$ & 1868 \\
		Trans=Yes & $0.5322$ & 31 \\
		\rowcolor{gray!15} ReligionImportant=No & $-0.1069$ & 1069 \\
		SO=Bisexual & $0.1639$ & 283 \\
		ToxProblem=Never & $0.2975$ & 92 \\
		\bottomrule
	\end{tabularx}
	\caption{Statistically significant ($p<0.05$) apunim results across all four datasets. Gray rows show negative values. Consult App.~\ref{app:full} for the full results.}
	\label{tab:merged-apunim}
\end{table}

\paragraph{Ethnicity} Across all datasets apart from Sap, we observe a consistent pattern in which annotation polarization is linked to annotator ethnicity, and particularly by \ac{aa} annotators. In fact, \emph{ethnicity is the strongest contributor to polarization} in the DICES datastes (\acp{aa}: $-0.0428$/Asians: $0.0659$ for DICES-350, all ethnicities for DICES-990) and the second strongest for Kumar (\acp{aa}: $0.2072$). These results align with prior work suggesting that ethnic background is a systematic source of annotator disagreement in these tasks~\cite{goyal2022your}.

\paragraph{Gender} Polarization relates to gender across both the Kumar and DICES-990 datasets. DICES-990 shows that men and women have statistically significant, opposite effects (Men: $0.028$; Women: $-0.025$). This pattern is similar to Sap, where men display a positive attribution score ($0.091$) and women display a strongly negative score ($-0.244$). These results are consistent with previous findings regarding hate speech~\citep{wojatzki2018women} and toxicity annotation~\cite{binns2017like}. Interestingly, the Kumar dataset--the only one providing this demographic data--indicates that transgender annotators showed the highest degree of polarization, even given their small representation in the dataset. This finding strongly supports our choice to use polarization as a metric (Fig.~\ref{fig:disagreement-vs-polarization}).

\begin{finding}
	\label{finding:categorical}
	Annotator ethnicity and gender are consistent causes of polarization, even when their representation in the annotations is sparse.
\end{finding}

\begin{figure}[t]
	\includegraphics[width=\linewidth]{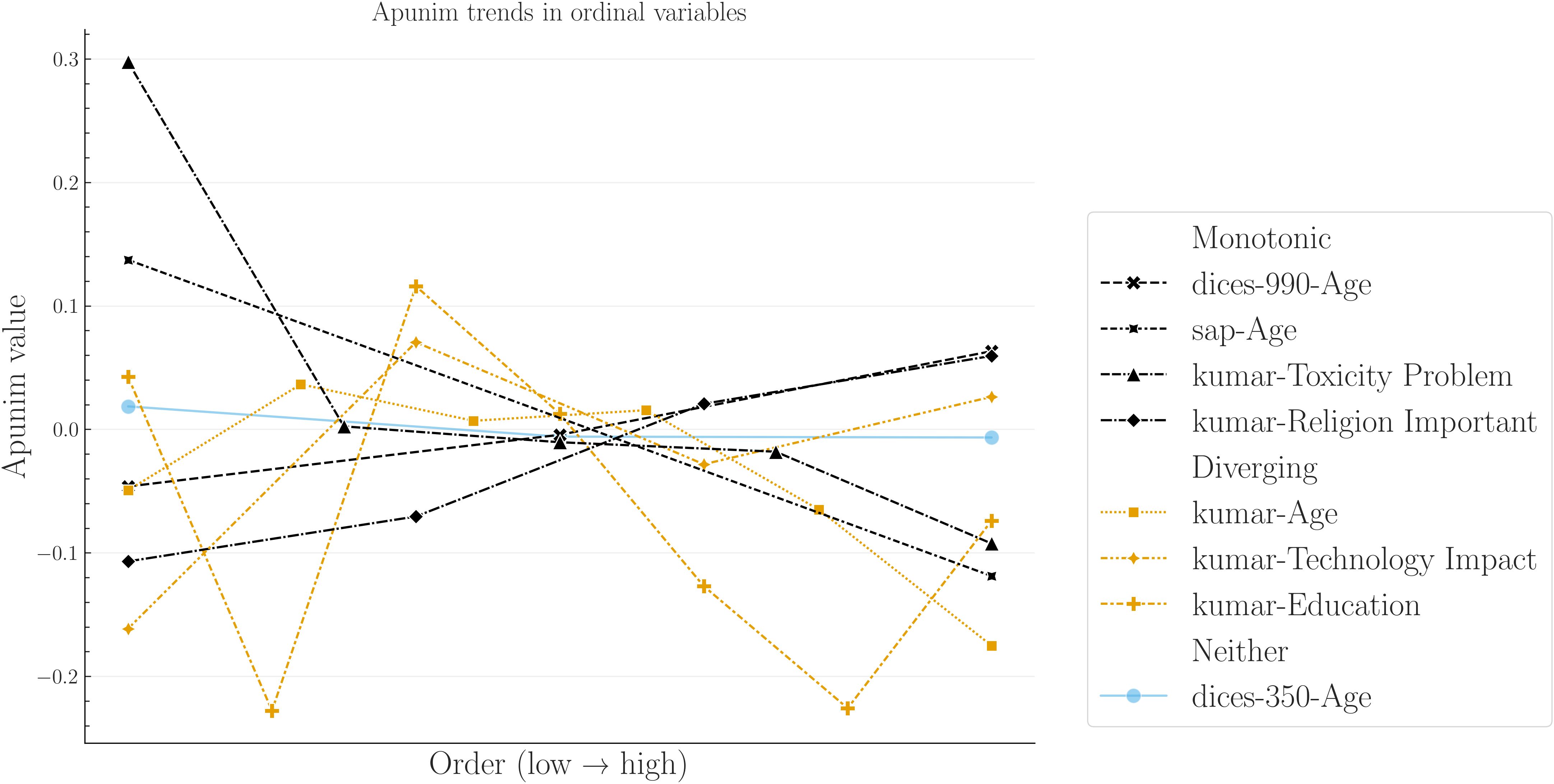}
	\caption{Polarization attribution trends in ordinal groups. The left side of the x-axis (low orders) refers to young annotators, low education, low religiousness, and negativity towards the impacts of toxicity and technology. The full ordinal modes for each dimension and their individual labels can be found in App.~\ref{app:full}.}
	\label{fig:ordinal}
\end{figure}

\paragraph{Trends}
Figure~\ref{fig:ordinal} shows \emph{apunim} trends across ordinal sociodemographic attributes listed in Table~\ref{tab:merged-apunim}. We find that \emph{polarization attribution changes consistently} as we move from one end of an attribute scale to the other--e.g., from irreligious to very religious annotators. 

In some sociodemographic dimensions (opinions on toxicity and religion) this is expressed in a monotonic pattern (black lines), suggesting that perceived differences become stronger between groups that are further apart ideologically or demographically. In other cases (opinions on technology and educational level), polarization attribution is caused by a few dominant modes (yellow lines). Finally, age seems to have a different effect depending on dataset.

\begin{finding}
	\label{finding:ordinal}
	For ordinal sociodemographic dimensions, polarization attribution tends either to grow stronger as the differences between annotator groups increase, or coalesce around a few opposing groups.
\end{finding}

\section{Conclusion \& Future work}

We investigated whether fundamental differences can be attributed to annotator groups using polarization, instead of standard agreement measures. We discovered as of yet unknown issues regarding the presence of inherent polarization and conflicting polarization directions which we bypass using a new metric, \textit{apunim}. We then applied our proposed metric to four hate-speech and toxicity detection datasets, discovering that (1) race and gender seem to generally drive polarization attribution, and (2) groups that are further apart increasingly drive polarization showing that even in challenging cases with very low annotator counts, \textit{apunim} is capable of attributing polarization trends. Finally, we estimated the robustness of our metric w.r.t. the number of annotators, and released a PyPi installable library implementing it.

Our framework has broad applications across \ac{nlp} tasks related to content moderation such as analyzing the evolution of polarization on the discussion level. Additionally, while we study one group at a time, we can instead turn to iteratively exploring polarized subgroups, to gradually piece together a multidimensional explanation of polarization (see Footnote~\ref{foot:iterative},~\S\ref{sec:inherent:formulation}). Finally, it would be interesting to explore whether LLM-as-a-judge systems reproduce the polarization patterns observed among different minority groups--which would solve the issues with annotation counts presented in our study.

\section*{Limitations}

Researchers often release only aggregated single-label annotations or omit annotator sociodemographic characteristics entirely. As a result, both our analysis and the applicability of our framework are limited by the scarcity of suitable datasets. In addition, because datasets with large numbers of annotators per item are rare, our approach operates only at the dataset level. This limitation prevents us from identifying individual comments or examples that could serve as concrete explanations for the observed \textit{apunim} values. 

Finally, although we compare findings across datasets and relate them to prior work, we do not directly compare polarization and agreement metrics, since they rely on fundamentally different assumptions and mechanisms (see \S\ref{sec:related:polarization}). These differences also make it difficult to fully explain certain patterns in our results, such as why some groups show statistically significant effects while others do not.

\section*{Ethical Considerations}

All datasets used in this study were either publicly accessible, or were provided to us by the authors with full disclosure on the purpose of our research. No identifiable, personal information is included in our pipeline. We did not perform any annotation tasks ourselves.
Additionally, this study uses dichotomies across the two predominant genders, and religious groups as examples. The authors do not claim that these examples necessarily reflect the views or behaviors of these groups; they exist only for the purpose of demonstration. The controversial statements expressed in this publication also do not express the views of the authors.

\section*{Acknowledgments}

This work has been partially supported by project MIS 5154714 of the National Recovery and Resilience Plan Greece 2.0 funded by the European Union under the NextGenerationEU Program.

We used LLMs for code generation and internal documentation in some instances. All outputs have been verified by the authors.

\bibliography{refs}
\appendix

\section {Acronyms}

\begin{acronym}[WWW]
	\acro{ndfu}[nDFU]{normalized Distance From Unimodality}
	\acro{fwer}[FWER]{Family-Wise Error Rate}
	\acro{nlp}[NLP]{Natural Language Processing}
	\acro{gpl}[GPLv3]{GNU General Public Licence v3}
	\acro{aa}[AA]{African American}
\end{acronym}

\section{Annotation details}
\label{sec:app:ann_details}

\begin{figure*}[t]
	\includegraphics[width=\columnwidth]{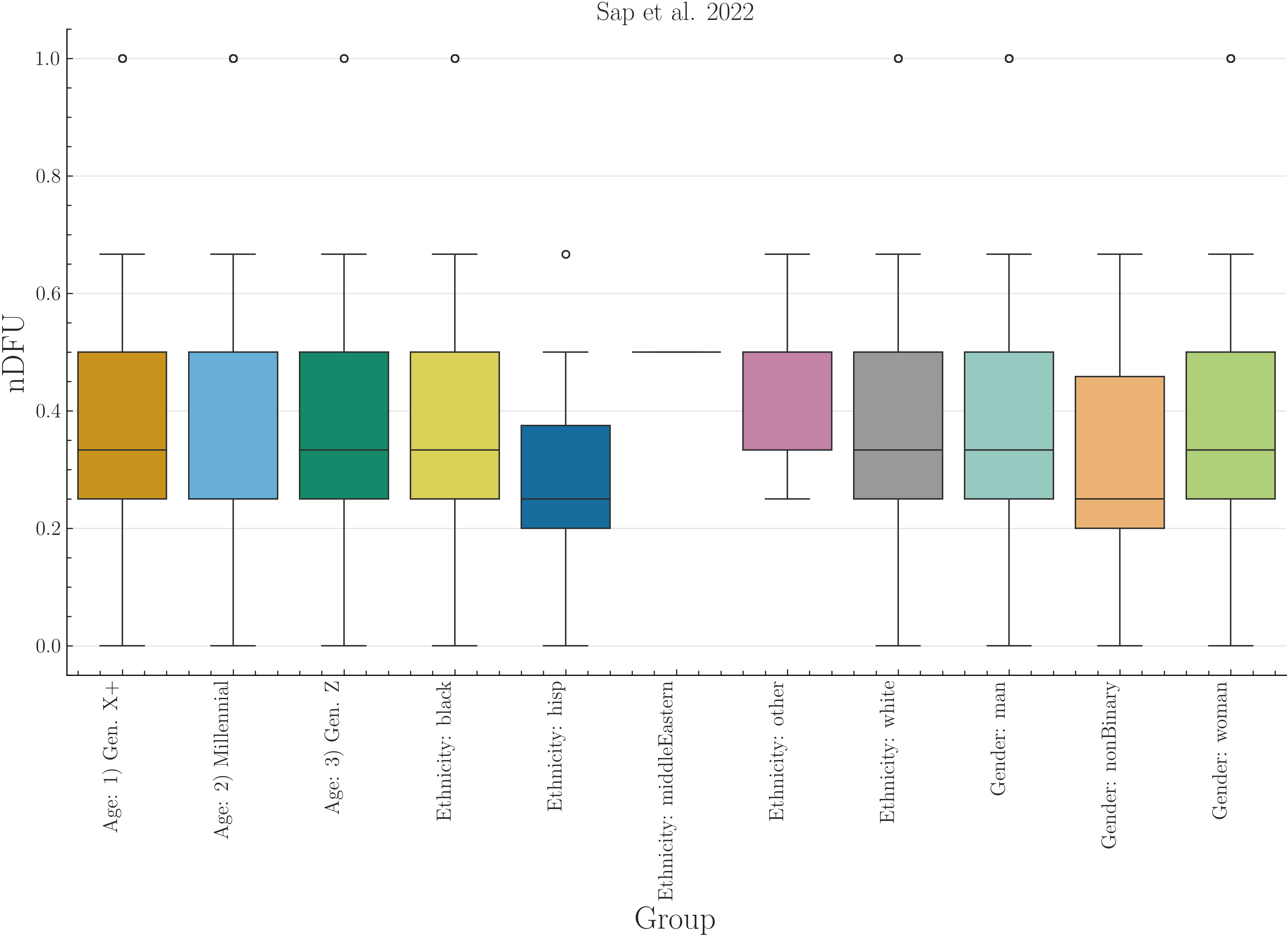}
	\includegraphics[width=\columnwidth]{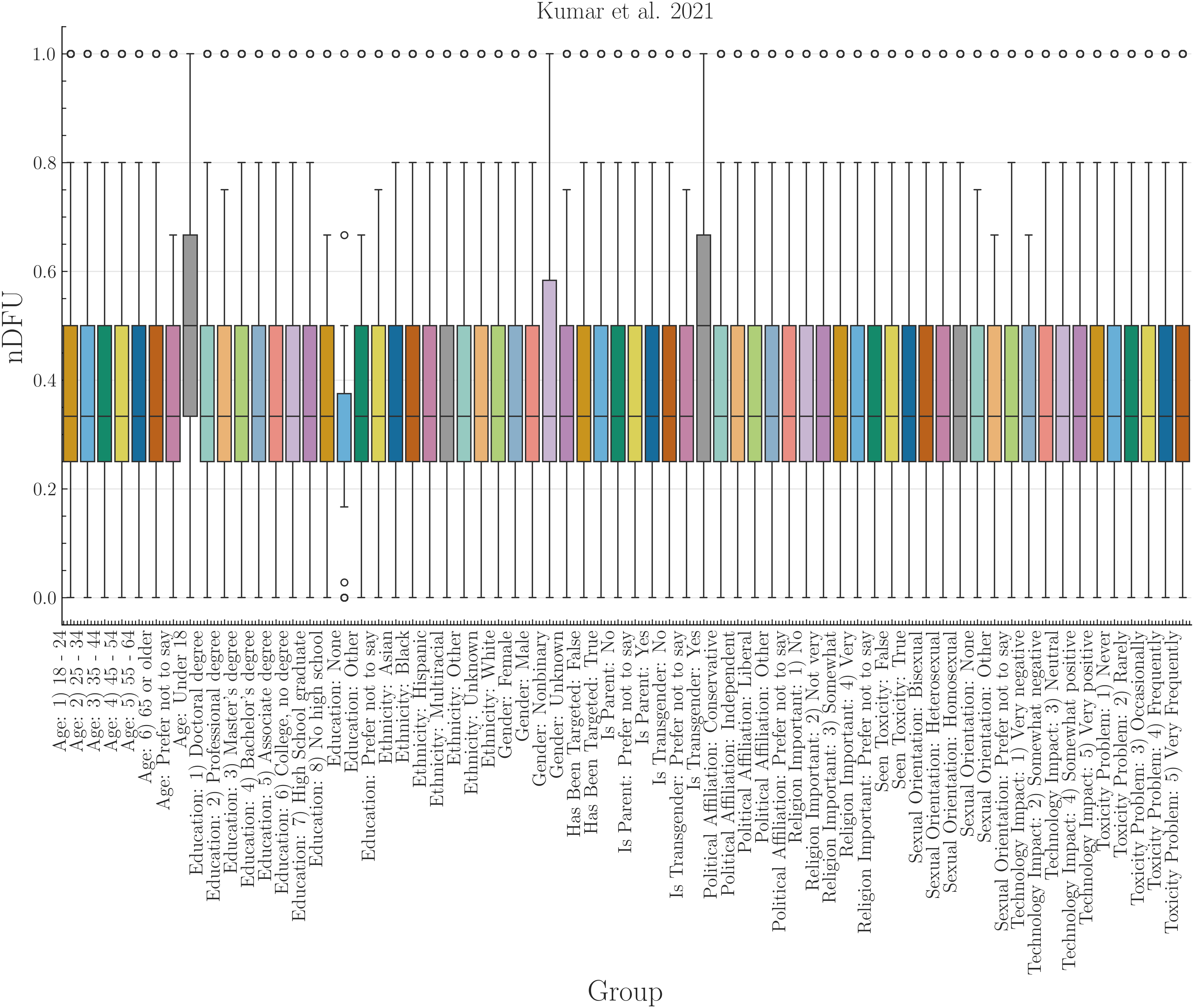}
	\includegraphics[width=\columnwidth]{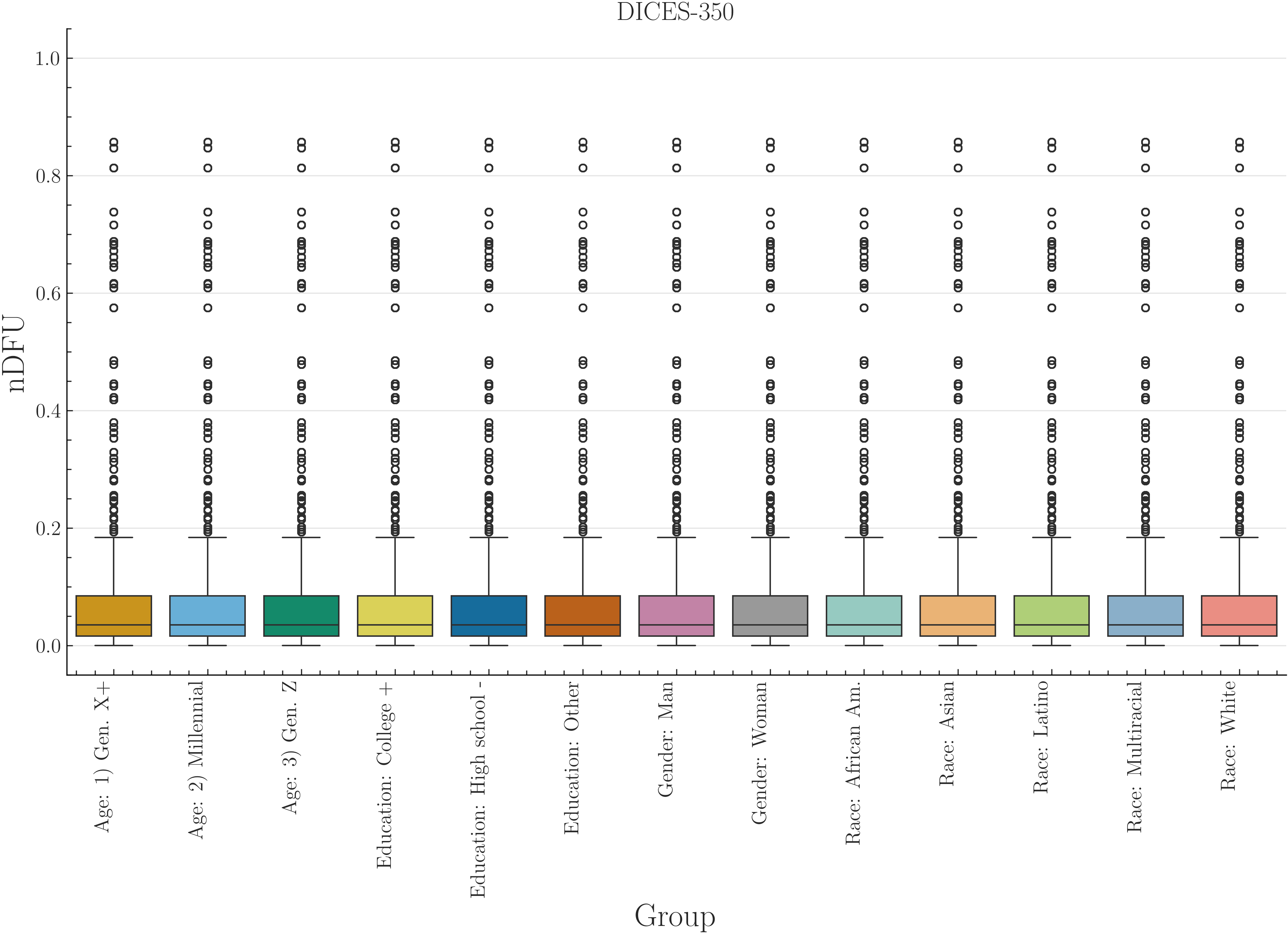}
	\includegraphics[width=\columnwidth]{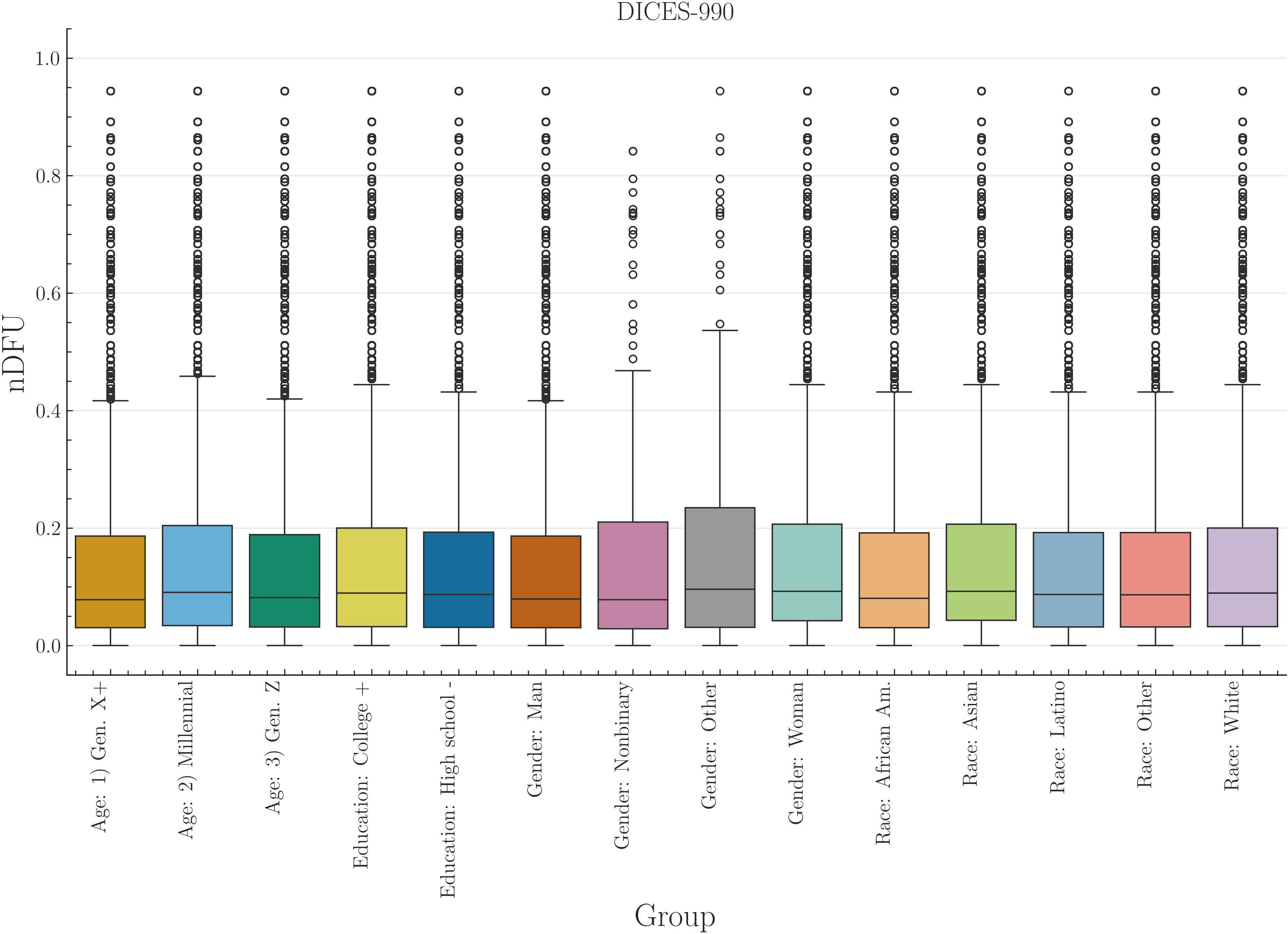}
	\caption{Histogram of dataset polarization for the human datasets considered in this paper per sociodemographic dimension.}
	\label{fig:dataset-polarization}
\end{figure*}

\begin{figure}[t]
	\includegraphics[width=\linewidth]{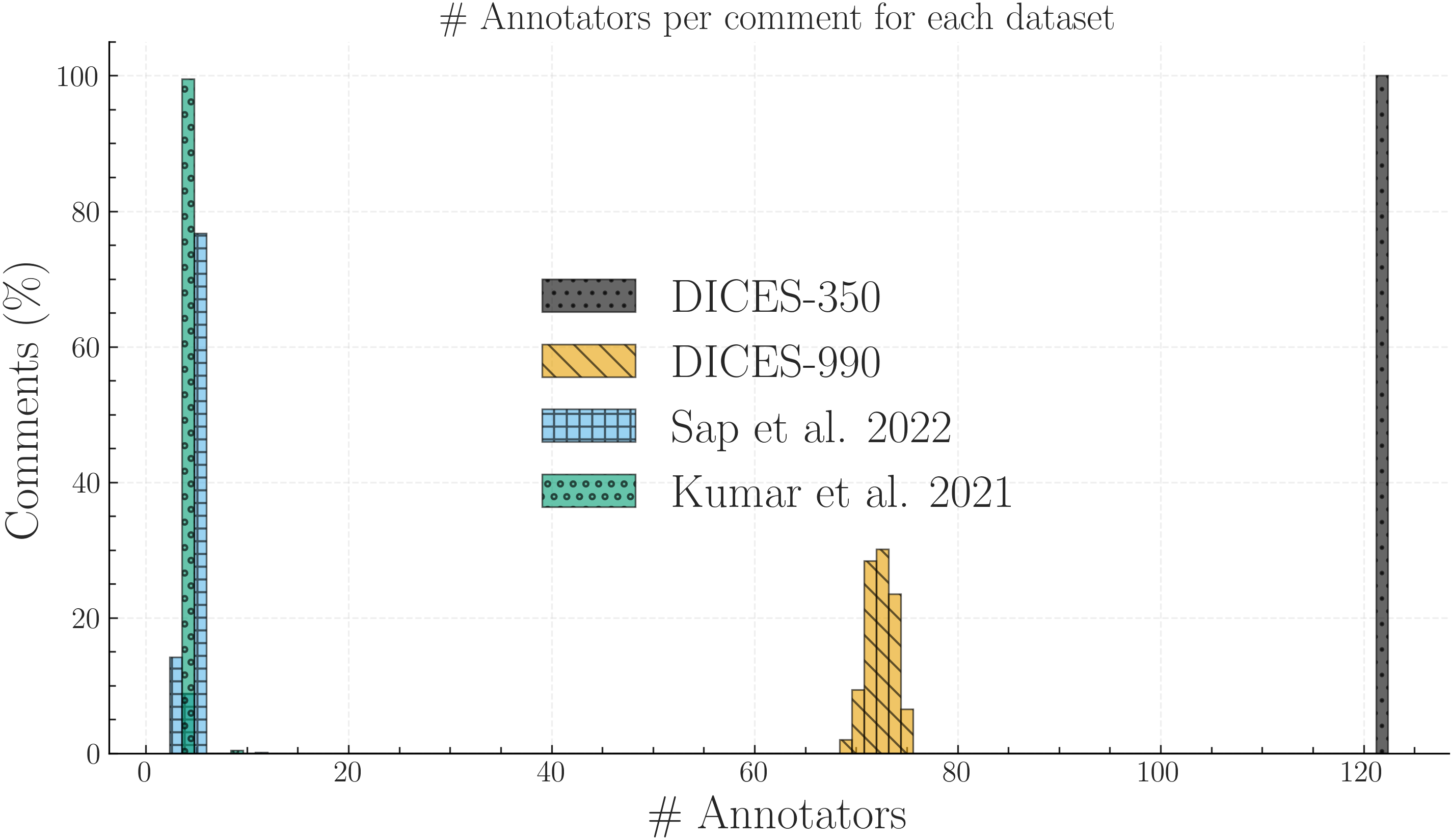}
	\caption{We calculate the ratio of comments for each dataset (y-axis) that had $x$ number of annotators (x-axis). We truncate the x-axis to 130 annotators, disregarding very few comments in the Kumar dataset exceeding this threshold (see Table~\ref{tab:num-annot}), which is likely caused by a data error.}
	\label{fig:ann_counts}
\end{figure}

\begin{table*}[t]
	\centering
	\caption{Descriptive statistics for the number of annotations per comment, grouped by dataset.}
	\label{tab:num-annot}
	\begin{tabular}{lrrrrrrrr}
		\toprule
		& count & mean & std & min & 25\% & 50\% & 75\% & max \\
		\midrule
		DICES-350 & 350 & 123 & 0 & 123 & 123 & 123 & 123 & 123 \\
		DICES-990 & 990 & 72.8313 & 1.1652 & 69 & 72 & 73 & 74 & 76 \\
		Kumar et al. 2021 & 20000 & 5.0925 & 4.8872 & 5 & 5 & 5 & 5 & 650 \\
		Sap et al. 2022 & 585 & 5.6359 & 0.7713 & 3 & 6 & 6 & 6 & 12 \\
		\bottomrule
	\end{tabular}
\end{table*}

The number of annotators per comment for all datasets presented in \S\ref{sec:inherent:datasets} can be found in Figure~\ref{fig:ann_counts}, and descriptive statistics can be found in Table~\ref{tab:num-annot}. The DICES-990 dataset has exactly 123 annotations per comment, while the other datasets exhibit a few variations around the central mean reported in Table~\ref{tab:datasets}. Interestingly, the sample used from the Kumar dataset exhibits a few comments that vastly exceed the promised $5$ number of annotators. We included these comments for the experiment presented in Table~\ref{tab:merged-apunim}, but not for the annotator number experiment in Figure~\ref{fig:std_error} since they are very likely a data error.

\section{P-value estimation}
\label{app:pvalue}

Since our test is parametric (\S\ref{sec:apunim:statistical}), it assumes that there exist (1) enough data-points to reliably run the means-test, (2) enough annotations per sociodemographic group to estimate \acp{ndfu} in Equations~\ref{eq:apriori}-\ref{eq:observed}, (3) the apunims of the random partitions are independent and identically distributed. Condition (1) can be safely assumed for most datasets, because they feature enough items to reliably assume normal residuals (see \S\ref{sec:inherent:datasets}). We discuss condition (2) in \S\ref{sec:apunim:ann-size}. Condition (3) is partly fulfilled; the partitions are identically distributed, but may be dependent on the sample distribution. 

Replacing the parametric means test with a non-parametric alternative would relax assumptions about the underlying distribution but would not resolve the potential dependence between the random apunims. We initially attempted to implement a permutation test on the empirical (random apunim) distribution, but this resulted in excessive computational costs.

While the test operates on a single group in principle, in practice we often want to test polarization for all subgroups of a specific dimension. Since we are simultaneously considering multiple hypotheses, we apply a multiple comparison correction to the resulting p-values--specifically the Holms method~\cite{holms}. The test is parameterized by the \ac{fwer}, which is used to tune the strength of the correction; we can increase this value to make our test more conservative towards multiple hypotheses~\cite{ChenFengYi2017}). In general, it is safe to set \ac{fwer} equal to the significance level of our test (e.g., $\textit{FWER} = 0.95$ if we are looking for $p < 0.05$).

\section{Software}
\label{app:software}

\subsection{Installation and usage}
We release the \texttt{apunim} package, which is a Python library that implements both the \ac{ndfu} and apunim calculations, as well as its statistical test. The software is available on PyPi We also provide high-level documentation in HTML format that includes an introduction, guide, and full low-level documentation for each of the two provided functions (\doclink).

The code is open-source, licensed under the \ac{gpl} and can be found on the project's repository (\explink). The library is cross-platform, works for any Python 3 version, and requires very few dependencies (namely \texttt{numpy, scipy, statsmodels}).

\subsection{Implementation details}
By default, the computation of the apunim values happens on a single thread. In cases of serious computational cost, we recommend running the apunim calculation on multiple concurrent processes, since all operations are CPU-bound and thread-safe. This may be useful in cases where there are multiple sociodemographic dimensions, each of which is computationally expensive. 

In order to lower the computational cost required, we re-use the random partitions $\tilde{r}_i$), which are originally calculated to obtain the apriori polarization values (see Equation~\ref{eq:apriori}), for the p-value calculations (\S\ref{sec:apunim:statistical}). This means that in our current implementation, the number of groups for estimating apriori polarization ($\tilde{r}_i$ in Equation~\ref{eq:apriori}), and the number of apunim values used in the p-value calculation have to be the same.

\subsection{Replication Details}
\label{app:replication}

We use version \texttt{1.0.2} of the \texttt{apunim} library for our experiments. We use 100 random iterations for the apriori value calculation (Equation~\ref{eq:apriori}) and the p-value estimation (\S\ref{sec:apunim:statistical}), and set a \ac{fwer} rate of $0.95$. 

While the library efficiently handles a large number of annotators through vectorized operations, performance degrades significantly when processing a high volume of comments, particularly when annotators belong to multiple distinct groups. For instance, while experiments across all datasets in \S\ref{sec:inherent:datasets} complete within a minute, the Kumar dataset necessitates approximately one hour of computation, even when using a 60\% sample. Furthermore, the annotator variance analysis detailed in \S\ref{sec:apunim:ann-size} required over ten hours. While this latter time investment is anticipated due to the inherently demanding nature and non-standard application of our library, it highlights a scaling bottleneck under specific data structures. We hypothesize that this bottleneck is likely caused by repeated creation of numpy arrays, resulting in repeated memory allocations for each subgroup, in each comment, for each sociodemographic dimension. We plan on profiling and addressing it in the next releases of our software. Until then, we suggest either using multiprocessing for each sociodemographic subgroup, or running the different experiments in parallel (indeed, we use the \texttt{gnu-parallel} linux package~\cite{tange_2025_16944306} for our own experiments).

In Figure~\ref{fig:std_error}, we removed values where there was a lack of items annotated by the specified number of annotators (e.g., only 7\% of comments in DICES-990 were annotated by more than 73 annotators---see App.~\ref{sec:app:ann_details}).

\section{Full Results}
\label{app:full}

\begin{table}[ht]
	\centering
	\small 
	\begin{tabularx}{\columnwidth}{X r r}
		\toprule
		\textbf{Group} & \textbf{apunim} & \textbf{support} \\
		\midrule
		\multicolumn{3}{c}{\textbf{Sap}} \\
		\midrule
		Age=GenX+ & $0.0660$ & 271 \\
		Age=Millennial & $-0.0068$ & 1886 \\
		\rowcolor{gray!15} Ethnicity=white & $0.0004$ & 1896 \\
		Gender=Man & $0.0907^{***}$ & 1439 \\
		\rowcolor{gray!15} Gender=Woman & $-0.2442^{***}$ & 877 \\
		\bottomrule
	\end{tabularx}
	\caption{Apunim results for the Sap Dataset. $^{*}$ denotes $p<0.10$, $^{**}$ denotes $p<0.05$, and $^{***}$ denotes $p<0.01$.  Rows shaded in gray indicate a negative $\text{apunim}$ value. Groups for which no polarized comments among groups not included.}
	\label{tab:sap-full}
\end{table}

\begin{figure*}[t]
	\includegraphics[width=\columnwidth]{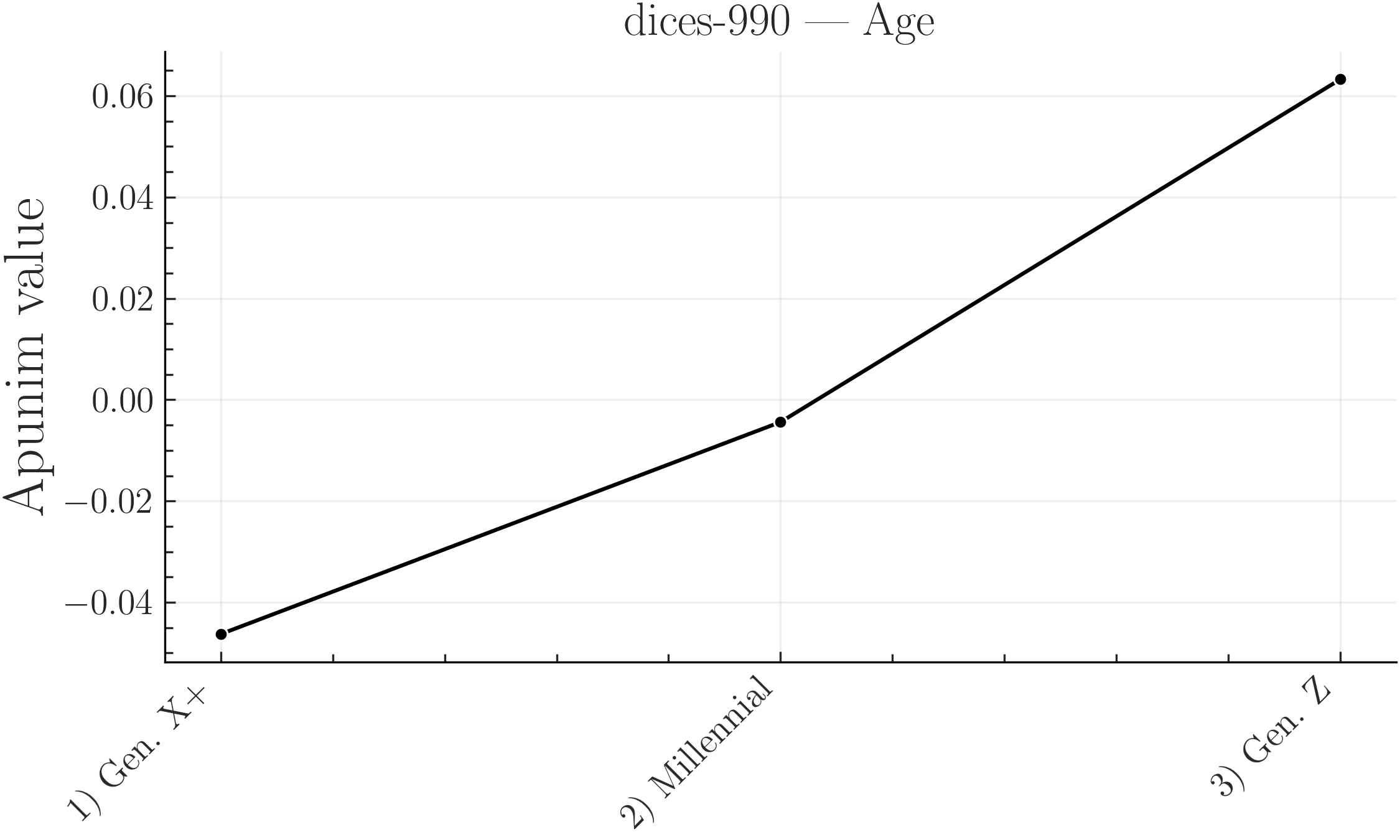}
	\includegraphics[width=\columnwidth]{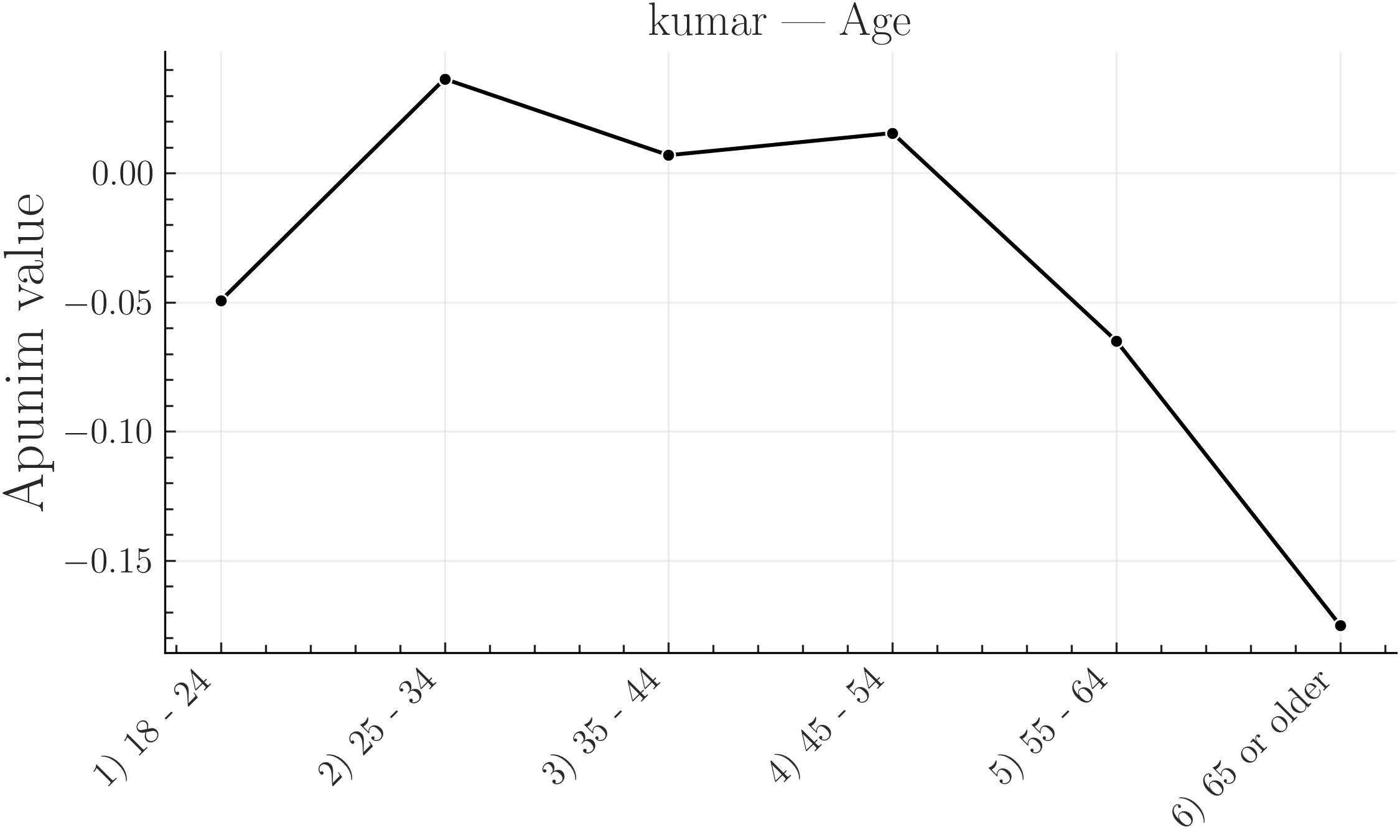}
	\includegraphics[width=\columnwidth]{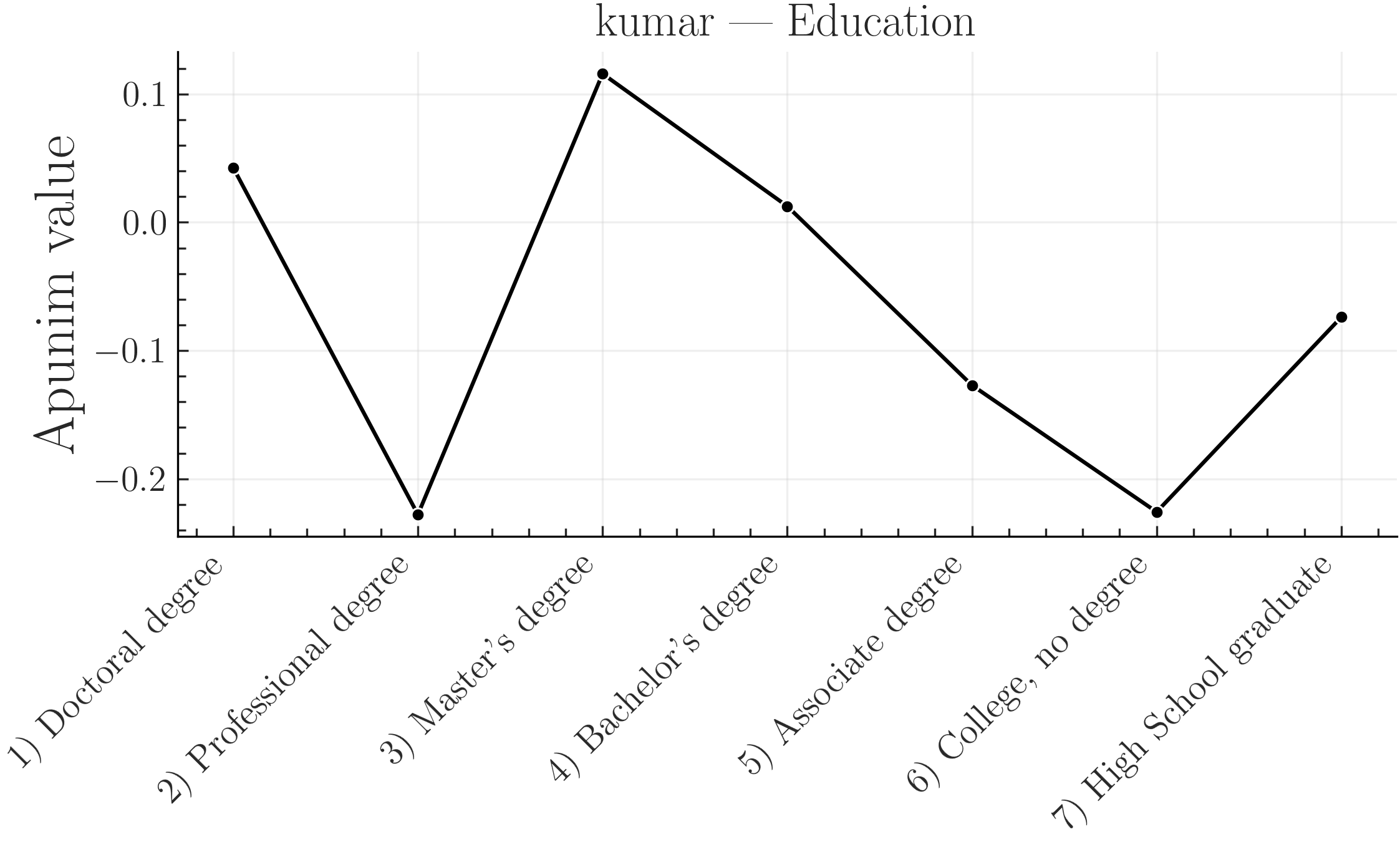}
	\includegraphics[width=\columnwidth]{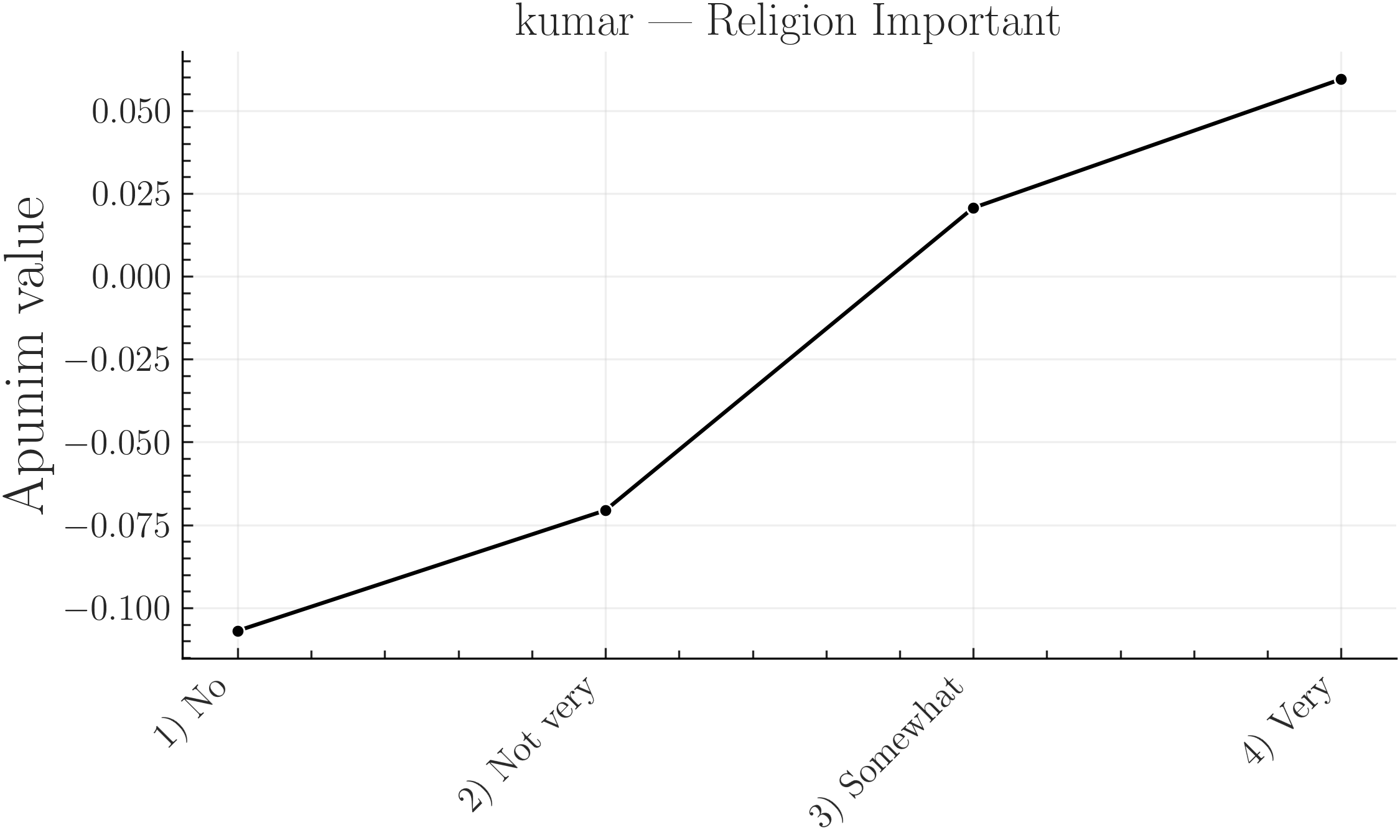}
	\includegraphics[width=\columnwidth]{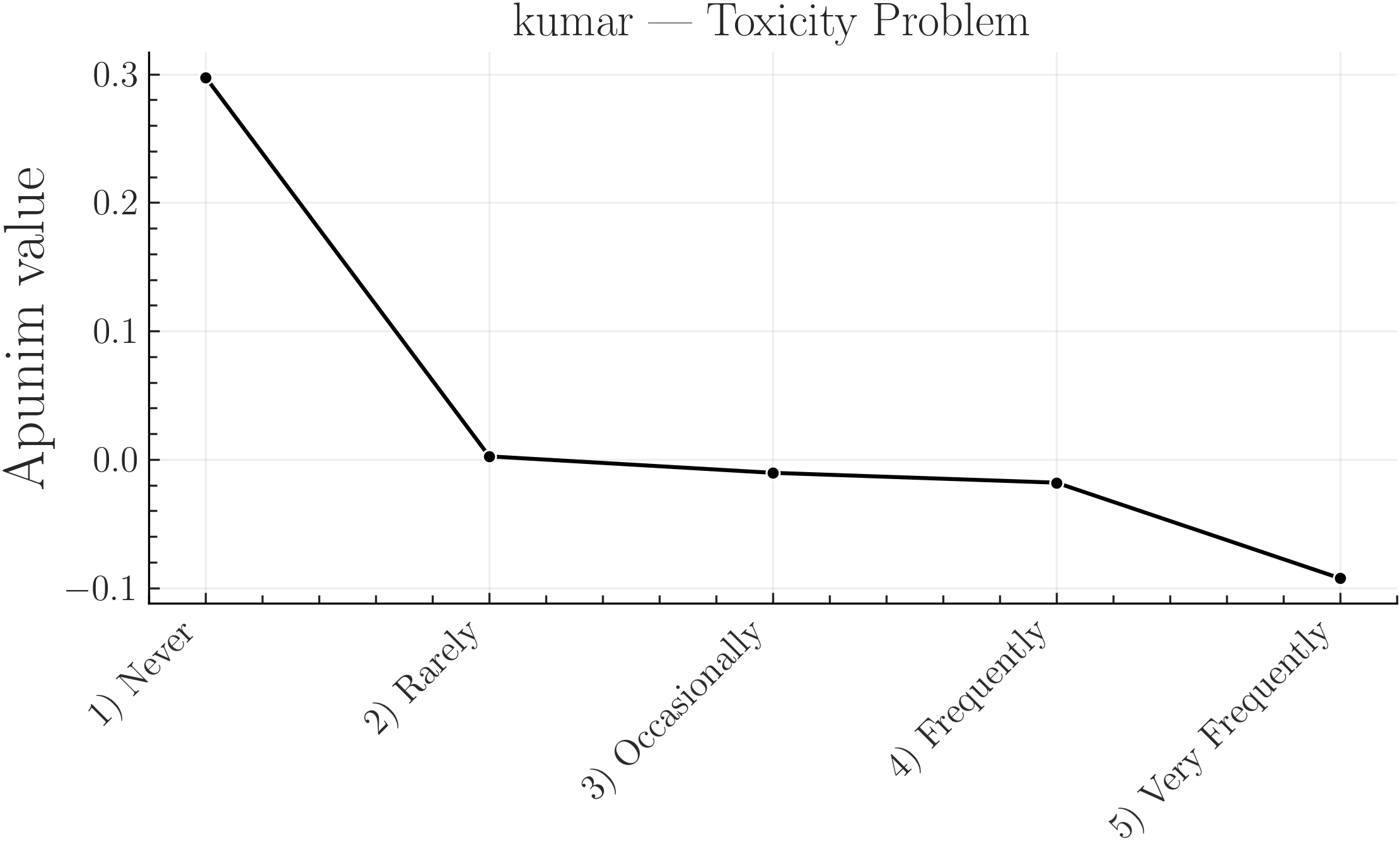}
	\caption{Apunim values across ordinal factor levels with at least two statistically significant values, expanded for each ordinal sociodemographic dimension. This figure is complementary to Figure~\ref{fig:ordinal}. The values in the graphs were extracted from the Tables of \ref{app:full}. }
	\label{fig:ordinal-full}
\end{figure*}

\begin{table}[t]
	\centering
	\small 
	\begin{tabularx}{\columnwidth}{X r r}
		\toprule
		\textbf{Group} & \textbf{apunim} & \textbf{support} \\
		\midrule
		\multicolumn{3}{c}{\textbf{Kumar}} \\
		\midrule
		\rowcolor{gray!15} Age=18-24 & $-0.0494$ & 223 \\
		Age=25-34 & $0.0365$ & 1188 \\
		Age=35-44 & $0.0069$ & 762 \\
		Age=45-54 & $0.0155$ & 291 \\
		\rowcolor{gray!15} Age=55-64 & $-0.0650$ & 130 \\
		\rowcolor{gray!15} Age=65+ & $-0.1752$ & 17 \\
		\midrule
		Edu=Doctoral degree & $0.0426$ & 8 \\
		\rowcolor{gray!15} Edu=Professional degree & $-0.2278$ & 7 \\
		Edu=Master's degree & $0.1159^{*}$ & 459 \\
		Edu=Bachelor's degree & $0.0126$ & 1193 \\
		\rowcolor{gray!15} Edu=Associate degree & $-0.1270$ & 176 \\
		\rowcolor{gray!15} Edu=College, no degree & $-0.2258^{***}$ & 540 \\
		\rowcolor{gray!15} Edu=High School grad & $-0.0739$ & 122 \\
		\midrule
		Asian & $0.0729$ & 75 \\
		Black & $0.2072^{***}$ & 356 \\
		Hispanic & $0.0255$ & 25 \\
		\rowcolor{gray!15} Multiracial & $-0.1099$ & 44 \\
		Other & $0.1992$ & 17 \\
		\rowcolor{gray!15} Unknown & $-0.0246$ & 7 \\
		\rowcolor{gray!15} White & $-0.0436$ & 1253 \\
		\midrule
		\rowcolor{gray!15} Gender=Female & $-0.0370$ & 2070 \\
		Gender=Male & $0.0422$ & 1974 \\
		\midrule
		\rowcolor{gray!15} Target=False & $-0.0017$ & 2199 \\
		Target=True & $0.0329$ & 1176 \\
		\midrule
		\rowcolor{gray!15} Parent=No & $-0.0591^{***}$ & 1868 \\
		\rowcolor{gray!15} Parent=Prefer not to say & $-0.1858$ & 4 \\
		Parent=Yes & $0.0205$ & 2156 \\
		\midrule
		\rowcolor{gray!15} Trans=No & $-0.0143$ & 561 \\
		Trans=Prefer not to say & $0.3697$ & 7 \\
		Trans=Yes & $0.5322^{**}$ & 31 \\
		\midrule
		Conservative & $0.0552$ & 1021 \\
		\rowcolor{gray!15} Independent & $-0.0610$ & 895 \\
		\rowcolor{gray!15} Liberal & $-0.0419$ & 1449 \\
		\rowcolor{gray!15} Other & $-0.1127$ & 21 \\
		\rowcolor{gray!15} Prefer not to say & $-0.0148$ & 54 \\
		\midrule
		\rowcolor{gray!15} Rel=No & $-0.1069^{***}$ & 1069 \\
		\rowcolor{gray!15} Rel=Not very & $-0.0706$ & 241 \\
		Somewhat & $0.0206$ & 837 \\
		Very & $0.0595$ & 1199 \\
		\rowcolor{gray!15} Prefer not to say & $-0.2144$ & 13 \\
		\midrule
		False & $0.0594$ & 849 \\
		\rowcolor{gray!15} True & $-0.0387$ & 2051 \\
		\midrule
		
		SO=Bisexual & $0.1639^{**}$ & 283 \\
		\rowcolor{gray!15} SO=Heterosexual & $-0.0353$ & 1170 \\
		\rowcolor{gray!15} SO=Homosexual & $-0.0103$ & 32 \\
		\rowcolor{gray!15} SO=Prefer not to say & $-0.2791$ & 23 \\
		\midrule
		
		Tech=Very negative & $0.2157$ & 4 \\
		\rowcolor{gray!15} Tech=Negative & $-0.1614$ & 176 \\
		Tech=Neutral & $0.0705$ & 320 \\
		\rowcolor{gray!15} Tech=Positive & $-0.0283$ & 1606 \\
		Tech=Very Positive & $0.0263$ & 1000 \\
		\midrule
		
		ToxProblem=Never & $0.2975^{**}$ & 92 \\
		ToxProblem=Rarely & $0.0024$ & 498 \\
		\rowcolor{gray!15} ToxProblem=Occasionally & $-0.0105$ & 1096 \\
		\rowcolor{gray!15}ToxProblem=Frequently & $-0.0181$ & 958 \\
		\rowcolor{gray!15} ToxProblem=Very Freq & $-0.0925$ & 252 \\
		\bottomrule
	\end{tabularx}
	\caption{Apunim results for the dataset. $^{*}$ denotes $p<0.10$, $^{**}$ denotes $p<0.05$, and $^{***}$ denotes $p<0.01$. Rows shaded in gray indicate a negative $\text{apunim}$ value.}
	\label{tab:sdb-features}
\end{table}

\begin{table}[t]
	\centering
	\small 
	\begin{tabularx}{\columnwidth}{X r r}
		\toprule
		\textbf{Group} & \textbf{apunim} & \textbf{support} \\
		\midrule
		\multicolumn{3}{c}{\textbf{DICES-350}} \\
		\midrule
		\rowcolor{gray!15} Race=AfricanAm. & $-0.0428^{***}$ & 9077 \\
		Race=Asian & $0.0659^{***}$ & 8138 \\
		\midrule
		\rowcolor{gray!15} Age=GenX+ & $0.0186$ & 9703 \\
		\rowcolor{gray!15} Age=Millennial & $-0.0059$ & 11268 \\
		\rowcolor{gray!15} Age=GenZ & $-0.0066$ & 17528 \\
		\midrule
		Edu=College or higher & $0.0124$ & 23475 \\
		\rowcolor{gray!15} Edu=H. school or lower & $-0.0152$ & 12833 \\
		\rowcolor{gray!15} Edu=Other & $-0.0214^{*}$ & 2191 \\
		\midrule
		\rowcolor{gray!15} Gender=Man & $-0.0193$ & 19093 \\
		Gender=Woman & $0.0215$ & 19406 \\
		\midrule
		\rowcolor{gray!15} Race=Latino & $-0.0030$ & 6886 \\
		\rowcolor{gray!15} Race=Multiracial & $-0.0001$ & 5008 \\
		\rowcolor{gray!15} Race=White & $-0.0198$ & 9390 \\
		\bottomrule
	\end{tabularx}
	\caption{Apunim results for the DICES-350 Dataset. $^{*}$ denotes $p<0.10$, $^{**}$ denotes $p<0.05$, and $^{***}$ denotes $p<0.01$.  Rows shaded in gray indicate a negative $\text{apunim}$ value. Groups for which no polarized comments among groups not included.}
	\label{tab:dices350-full}
\end{table}

\begin{table}[t]
	\centering
	\small 
	\begin{tabularx}{\columnwidth}{X r r}
		\toprule
		\textbf{Group} & \textbf{apunim} & \textbf{support} \\
		\midrule
		\multicolumn{3}{c}{\textbf{DICES-990}} \\
		\midrule
		\rowcolor{gray!15} Age=GenX+ & $0.0186$ & 9703 \\
		\rowcolor{gray!15} Age=Millennial & $-0.0044$ & 38436 \\
		Age=GenZ & $0.0633^{***}$ & 13741 \\
		\midrule
		Edu=College or higher & $0.0094$ & 59142 \\
		\rowcolor{gray!15} Edu=H. school or lower & $-0.0752^{***}$ & 7419 \\
		Edu=Other & $-0.0214^{*}$ & 2191 \\
		\midrule
		Gender=Man & $0.0275^{***}$ & 32494 \\
		\rowcolor{gray!15} Gender=Woman & $-0.0245^{***}$ & 33471 \\
		\midrule
		Race=AfricanAm. & $0.0911^{***}$ & 5810 \\
		Race=Asian & $0.0107$ & 18428 \\
		\rowcolor{gray!15} Race=Latino & $-0.1331^{***}$ & 6610 \\
		\rowcolor{gray!15} Race=Multiracial & $-0.0001$ & 5008 \\
		\rowcolor{gray!15} Race=White & $-0.1197^{***}$ & 11392 \\
		\bottomrule
	\end{tabularx}
	\caption{Apunim results for the DICES-990 Dataset. $^{*}$ denotes $p<0.10$, $^{**}$ denotes $p<0.05$, and $^{***}$ denotes $p<0.01$.  Rows shaded in gray indicate a negative $\text{apunim}$ value. Groups for which no polarized comments among groups not included.}
	\label{tab:dices990-full}
\end{table}

\end{document}